\documentclass{article}
\usepackage{iclr2025_conference,times}

\usepackage{amsmath,amsfonts,bm}

\def\eqref#1{equation~\ref{#1}}

\def\1{\bm{1}}

\DeclareMathAlphabet{\mathsfit}{\encodingdefault}{\sfdefault}{m}{sl}
\SetMathAlphabet{\mathsfit}{bold}{\encodingdefault}{\sfdefault}{bx}{n}

\usepackage{amsmath}
\usepackage{amssymb}
\usepackage{natbib}
\usepackage{float}
\usepackage{tabu}
\usepackage{algorithm}
\RequirePackage{algorithmic}
\usepackage{fontawesome5}
\usepackage{hyperref}
\usepackage{url}
\usepackage[utf8]{inputenc} 
\usepackage{url}           
\usepackage{booktabs}       
\usepackage{amsfonts}       
\usepackage{nicefrac}       
\usepackage{microtype}      
\usepackage{xcolor}         
\usepackage{tcolorbox}

\usepackage{graphicx}
\usepackage{longtable}
\usepackage{caption}
\usepackage{mdframed}
\usepackage{subcaption}
\usepackage{multirow}
\usepackage{placeins}
\usepackage{multicol}
\usepackage{makecell}
\usepackage{soul}
\usepackage[normalem]{ulem} 
\usepackage{wrapfig}
\usepackage[percent]{overpic}
\usepackage{lipsum}
\usepackage{csquotes}
\usepackage[OT2,T1]{fontenc}
\usepackage[english]{babel}
\usepackage{devanagari}
\usepackage{tablefootnote}
\usepackage{pdfpages}
\usepackage{colortbl}
\usepackage{arydshln}
\usepackage{array}
\usepackage{enumitem}
\usepackage{capt-of}

\definecolor{textpurple}{RGB}{0,0,0}
\definecolor{myblue}{RGB}{0, 0, 255}
\definecolor{mygray}{gray}{0.6}
\definecolor{lightgray}{gray}{0.91}

\definecolor{upcolor}{RGB}{57,182,74}
\newcommand{\up}[1]{\textcolor{upcolor}{$\uparrow$ #1}}
\newcommand{\down}[1]{\textcolor{red}{$\downarrow$ #1}}

\newcommand{\myparagraph}[1]{{\bf #1}}

\usepackage{pifont}

\def\halfcheckmark{\textcolor{black}{\ding{52}}{\small\textcolor{black}{\kern-0.7em\ding{55}}}}

\newcommand{\authorskip}{\hspace{4.8mm}}

\definecolor{myGreen}{rgb}{0, .6, .0}
\definecolor{myorange}{RGB}{251, 122, 56}
\definecolor{goodblue}{HTML}{0071bc}
\hypersetup{colorlinks=true,breaklinks,linkcolor=red,citecolor=goodblue}

\usepackage{cleveref}
\crefname{section}{Sec.}{Secs.}
\Crefname{section}{Section}{Sections}
\crefname{appendix}{Appendix}{Appendixes}
\Crefname{appendix}{Appendix}{Appendixes}
\crefname{table}{Tab.}{Tabs.}
\Crefname{table}{Table}{Tables}
\crefname{figure}{Fig.}{Figs.}
\Crefname{figure}{Figure}{Figures}
\crefname{equation}{Eq.}{Eqs.}
\Crefname{equation}{Equation}{Equations}

\title{
Reconstructive Visual Instruction Tuning
}

\usepackage{xspace}
\def\method{{\sc\textbf{Ross}}\xspace}

\author{
Haochen Wang$^{1,2}$
\authorskip Anlin Zheng$^3$
\authorskip Yucheng Zhao$^{4\dag}$
\authorskip Tiancai Wang$^{4*}$
\\
~\textbf{Zheng Ge}$^5$ \authorskip \textbf{Xiangyu Zhang}$^{4,5}$
\authorskip \textbf{Zhaoxiang Zhang}$^{1,2,6}$\thanks{Corresponding authors. $\dag$ Project lead.}
\\[2mm]
$^1$ Institute of Automation, Chinese Academy of Sciences \\
$^2$ University of Chinese Academy of Sciences \\
$^3$ University of Hong Kong \authorskip $^4$ MEGVII Technology \authorskip $^5$ 
 StepFun \\
$^6$ Centre for Artificial Intelligence and Robotics, \\
\ \ \ Hong Kong Institute of Science \& Innovation, Chinese Academy of Science \\[1pt]
{\small\texttt{
\{wanghaochen2022, zhaoxiang.zhang\}@ia.ac.cn
}} \quad {\small\texttt{wangtiancai@megvii.com}}\\[2mm]
\centerline{Project Page: \small{\url{https://haochen-wang409.github.io/ross}}}
}

\iclrfinalcopy
\begin{document}

\maketitle

\newcommand{\bzero}{\mathbf{0}}
\newcommand{\bone}{\mathbf{1}}
\newcommand{\cond}{\textbf{c}}

\begin{abstract}
This paper introduces \underline{\textbf{r}}ec\underline{\textbf{o}}nstructive vi\underline{\textbf{s}}ual in\underline{\textbf{s}}truction tuning (\method), a family of Large Multimodal Models (LMMs) that exploit \textit{vision-centric supervision signals}.
In contrast to conventional visual instruction tuning approaches that exclusively supervise text outputs, \method prompts LMMs to supervise \textit{visual outputs} via reconstructing input images.
By doing so, it capitalizes on the inherent richness and detail present within input images themselves, which are often lost in pure text supervision.
However, producing meaningful feedback from natural images is challenging due to the heavy spatial redundancy of visual signals.
To address this issue, \method employs a denoising objective to reconstruct latent representations of input images, avoiding directly regressing exact raw RGB values.
This \textit{intrinsic activation} design inherently encourages LMMs to maintain image detail, thereby enhancing their fine-grained comprehension capabilities and reducing hallucinations.
Empirically, \method \textit{consistently} brings significant improvements across different visual encoders and language models.
In comparison with \textit{extrinsic assistance} state-of-the-art alternatives that aggregate multiple visual experts, \method delivers competitive performance with a single SigLIP visual encoder, demonstrating the efficacy of our vision-centric \textit{supervision} tailored for \textit{visual outputs}.
%
\end{abstract}
\section{Introduction}\label{sec:intro}
\vspace{-5pt}
The success of GPT-style Large Language Models (LLMs)~\citep{radford2018improving, radford2019language, brown2020language, openai2023gpt, yang2024qwen2, touvron2023llama2, vicuna2023, dubey2024llama} has motivated researchers to adapt LLMs to understand multimodal inputs~\citep{liu2023visual, liu2024improved, instructblip, bai2023qwen}.
Notably, visual instruction tuning approaches~\citep{liu2023visual} demonstrate superior performance with cost-efficient training recipes.
%
%
Some approaches~\citep{chen2024far, li2024llava} even surpass GPT-4V(ision)~\citep{gpt4v} on benchmark evaluations.

\begin{figure}[t]
    \centering
    \includegraphics[width=0.95\linewidth]{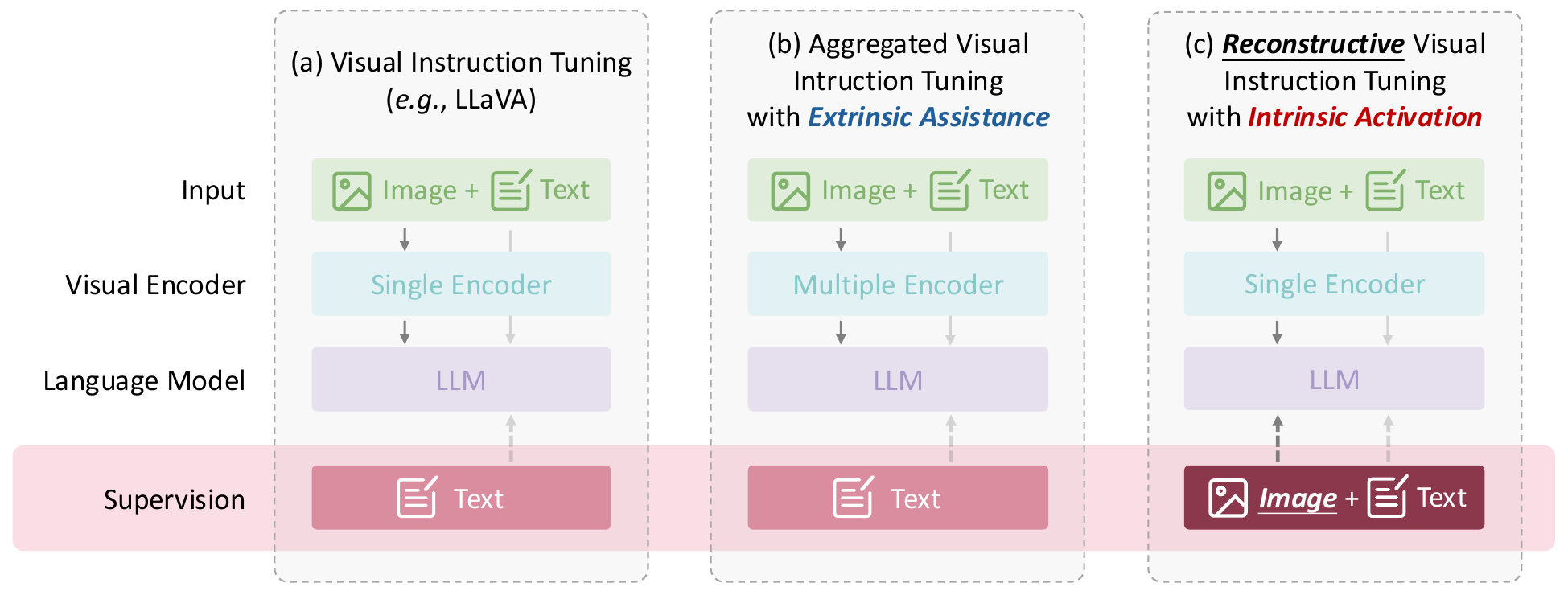}
    \vspace{-5pt}
    \caption{
    \textbf{Conceptual comparison} between different pipelines.
    \textbf{(a)} Typical visual instruction tuning approaches~\citep{liu2023visual, liu2024improved} follow a LLM-centric design that solely leverage text supervision.
    \textbf{(b)} Aggregated visual instruction tuning alternatives~\citep{tong2024cambrian, tong2024eyes} leverages \textit{extrinsic assistance} via combining several visual experts, requiring a careful selection of visual experts.
    \textbf{(c)} Our \method, with a single visual encoder, \textit{e.g.}, CLIP~\citep{radford2021learning} and SigLIP~\citep{zhai2023sigmoid}, designs extra vision-centric reconstructive supervision as \textit{intrinsic activation}.
    In this way, LMMs are required to preserve every detail of input images, thereby enhancing multimodal comprehension capabilities and reducing hallucinations.
    %
    }
    \label{fig:compare}
    \vspace{-10pt}
\end{figure}

Typically, these Large Multimodal Models (LMMs) based on visual instruction tuning adopt a plug-in architecture, as depicted in \Cref{fig:compare}\textcolor{red}{a}, where pre-trained vision-language foundation models such as CLIP~\citep{radford2021learning} are responsible for projecting images into visual tokens.
They serve as prefix tokens for multimodal comprehension.
However, this type of design, \textit{i.e.}, \textit{visual encoder $\to$ connector $\to$ LLM $\Leftarrow$ language instructions}, where ``$\Leftarrow$'' indicates supervision, is primarily LLM-centric: \textit{\textbf{(i)}} visual comprehension largely depends on vision-to-text alignment and the selected vision models, and \textbf{\textit{(ii)}} \textit{supervision} derives exclusively from text data.
As a result, they exhibit systematic visual shortcomings such as recognizing specific visual patterns~\citep{tong2024eyes}.

Until very recently, some concurrent works proposed \textit{vision-centric} solutions~\citep{tong2024cambrian, tong2024eyes}.
Illustrated in \Cref{fig:compare}\textcolor{red}{b}, their solutions leverage \textit{extrinsic assistance} via aggregating several different visual experts.
%
%
Inspired by the evolution in image recognition, from manually designed visual features~\citep{sanchez2011high} to learnable deep convolutional models~\citep{krizhevsky2012imagenet}, we suggest that \textit{intrinsic activation} offers a more viable path forward.
Just as deep models automatically learn hierarchical and abstract features from raw data, we believe \textit{intrinsic activation} methods are similarly more adaptable for multimodal comprehension, reducing reliance on hand-crafted engineering, thereby enhancing both generalization and performance.
Therefore, we aim to explore \textit{intrinsic activation} solutions  based on the following principles:
\begin{enumerate}
    \item \textbf{Supervise Visual Outputs.}
    Current LMMs solely supervise text outputs, neglecting a significant amount of visual outputs \textit{unused}.
    For instance, LLaVA-v1.5 \citep{liu2024improved} utilizes $576$ visual tokens to represent a single $336\times 336$ image, yet their corresponding outputs remain \textit{unsupervised}.
    Intuitively, since input images themselves inherently provide rich and detailed information, we regard LMMs \textit{reconstructing input images} as the supervision of those visual outputs.
    This approach encourages LMMs to maintain low-level details, thereby enhancing their fine-grained comprehension abilities and reducing hallucinations.
    \item \textbf{Explore the Optimal Formulation.}
    Designing this \textit{self-supervised} task effectively is not straightforward.
    Motivated by the success of \textit{masked} autoencoder~\citep{he2022masked} compared to its basic version denoising autoencoder~\citep{vincent2008extracting}, we identify handling \textit{heavy spatial redundancy of visual signals} as the underlying key factor.
    To this end, we formulate our approach as follows: \textbf{\textit{(i)}} for reconstruction \textit{targets}, instead of raw RGB pixels, we make LMMs reconstruct \textit{latent visual tokens}, and \textbf{\textit{(ii)}} for reconstruction \textit{objectives}, to avoid directly regressing exact token values, we adopt per-token \textit{denoising}.
\end{enumerate}
To this end, we propose \method, termed of \underline{\textbf{r}}ec\underline{\textbf{o}}nstructive vi\underline{\textbf{s}}ual in\underline{\textbf{s}}truction tuning, which utilizes input images as direct supervision signals illustrated in \Cref{fig:compare}\textcolor{red}{c}. 
Technically, to address the spatial redundancy inherent in natural visual signals~\citep{he2022masked}, we train a small denoising network, which takes high-level visual outputs $x$ as conditions to recover low-level fine-grained visual tokens $z$, representing an underlying distribution $p(z|x)$.
These latent tokens $z$ are derived from a frozen teacher tokenizer such as continuous VAE~\citep{kingma2013auto} and discrete VQGAN~\citep{esser2021taming}.
Unlike \textit{extrinsic assistance} solutions~\citep{tong2024cambrian, tong2024eyes}, our \textit{intrinsic activation} solution naturally maintains a lightweight inference procedure.
More importantly, when adapting to new visual domains, our solution avoids a careful choice of new domain-specific experts, \textit{e.g.}, MiDaS-3.0~\citep{birkl2023midas} for understanding depth maps, which is more efficient and easier to implement.

Empirically, \method achieves top performance across a wide range of multimodal comprehension benchmarks.
Notably, our \method excels in fine-grained vision-centric benchmarks~\citep{tong2024eyes, masry2022chartqa} and hallucination benchmarks~\citep{guan2024hallusionbench, li2023evaluating}.
%
%
To be specific, with a \textit{single} SigLIP~\citep{zhai2023sigmoid} as the visual encoder, \textbf{\method-7B} achieves 57.3 on HallusionBench~\citep{guan2024hallusionbench} and 54.7 on MMVP~\citep{tong2024eyes}, significantly outperforms state-of-the-art alternatives with similar model sizes which aggregate several visual experts as \textit{extrinsic assistance}, \textit{e.g.}, Cambrian-1-8B~\citep{tong2024cambrian}.
In-depth analysis demonstrates the effectiveness of \method for directing focus towards visual elements and understanding depth maps.
We hope our research will inspire future work in designing supervision signals for large multimodal models.

\section{Related Work}
\vspace{-5pt}
\myparagraph{Visual Instruction Tuning.}
Most visual instruction tuning-based LMMs adopt a plug-in architecture~\citep{liu2023visual, liu2024improved, bai2023qwen, wang2024world}, where a language-supervised visual encoder~\citep{radford2021learning, zhai2023sigmoid} is responsible for extracting visual tokens.
A connector is used to map those visual representations into the LLM space, \textit{e.g.}, Resamplers~\citep{alayrac2022flamingo}, Q-Formers~\citep{li2023blip2, instructblip, bai2023qwen, ge2024making}, and MLPs~\citep{liu2023visual, liu2024improved, li2024llava, liu2024llavanext, li2024llavanext-strong}.
These LMMs usually follow a two-stage training recipe.
During the alignment stage, the connector is trained on high-quality caption data.
Next, the full model is trained on single-image visual instruction tuning data.
However, \textit{only text outputs} are supervised.
\method, on the other hand, introduces novel vision-centric supervision via reconstructing fine-grained visual tokens conditioned on \textit{visual outputs}.
%

\myparagraph{Visual Encoders for LMMs.}
As the original CLIP~\citep{radford2021learning} adopted by conventional visual instruction tuning approaches is trained on noisy image-text pairs, it exhibits specific visual shortcomings, and thus stronger backbones~\citep{fang2024eva,zhai2023sigmoid,chen2024internvl} have been introduced to LMMs.
Some concurrent works~\citep{tong2024eyes, tong2024cambrian} leverage \textit{extrinsic assistance}, which further utilizes vision-only self-supervised models~\citep{oquab2023dinov2, wang2023bootstrap, wang2023droppos, wang2023hard, he2022masked, caron2021emerging} and domain experts~\citep{kirillov2023segment, birkl2023midas, rombach2022high}.
\method, from a new \textit{intrinsic activation} perspective, aims to catalyze enhanced comprehension through \textit{reconstructing input images} with \textit{no} extra visual experts.
%
%

\myparagraph{Generative Objectives for LMMs.}
Another line of work introduces \textit{pre-trained} text-to-image diffusion models~\citep{rombach2022high} to make LMMs capable of both comprehension and \textit{generation}~\citep{dong2024dreamllm, ge2024making, sun2024generative, ge2024seed, sun2023generative}.
Our \method, with a totally different motivation, targets to catalyze multimodal comprehension via \textit{reconstruction}.
%
%
Specifically, conditions are different, where \cite{dong2024dreamllm} and \cite{sun2024generative} take outputs corresponding to \textit{learnable queries} as conditions, while our \method takes outputs corresponding to \textit{visual inputs}.
Those methods are \textit{generative} while \method is \textit{reconstructive}.
The detailed pipeline comparison can be found in \Cref{sec:more_exp}.
%

\section{Preliminaries}\label{sec:pre}
\vspace{-5pt}
\myparagraph{Large Multimodal Models.}
In the literature~\citep{radford2018improving, radford2019language}, a $\theta$-parameterized LLM models the canonical \textit{causal} distribution of each \textit{text} token $\bm{x}_i$ as $p_{\theta}(\bm{x}) = \prod_{i=1}^T p_{\theta} (\bm{x}_i | \bm{x}_{<i})$, where $\{\bm{x}_i\}_{i=1}^T$ represents a sequence of text tokens.
To make LLMs understand visual contents, typical plug-in style LMMs~\citep{liu2023visual, liu2024improved} regard a sequence of visual tokens as prefix tokens.
Specifically, an input image $\bm{I} \in \mathbb{R}^{H \times W \times 3}$ is first projected into a sequence of visual tokens by a $\xi$-parameterized visual encoder $\mathcal{G}_{\xi}$ such as CLIP~\citep{radford2021learning} and SigLIP~\citep{zhai2023sigmoid}, where $(H, W)$ indicates the spatial resolution.
Then, a $\phi$-parameterized multimodal projector $\mathcal{H}_{\phi}$ is utilized to project these visual tokens into the feature space of LLMs.
As a result, the canonical causal distribution in a \textit{multimodal} sentence containing an image $\bm{I}$ becomes
\begin{equation}
    p_{\Theta} (\bm{x}) = \prod_{i=1}^T p_{\Theta} (\bm{x}_i | \bm{x}_{<i}, \bm{v}), \quad \bm{v} = \mathcal{H}_{\phi} \circ \mathcal{G}_{\xi} (\bm{I}),
\end{equation}
where $\Theta = \{\theta, \xi, \phi\}$ is the parameters and $\bm{v} \in \mathbb{R}^{N \times D}$ indicates the projected visual tokens.
$N$ is the number of visual tokens and $D$ indicates the feature channel.
The visual encoder $\mathcal{G}_{\xi}$ could be either frozen~\citep{liu2023visual, liu2024improved, tong2024cambrian} or fine-tuned~\citep{liu2024llavanext, bai2023qwen, li2024llava, wang2024qwen2}.

\myparagraph{Training Recipes for LMMs.}
LMMs almost follow a two-stage training recipe~\citep{liu2023visual}, \textit{i.e.}, the pre-training stage (or the alignment stage) and the supervised fine-tuning stage (or the instruction tuning stage).
The instruction (supervision) comes from languages such as the answers to VQA tasks, maximizing the log-likelihood of \textit{text} outputs:
\begin{equation}
\label{eq:lmm}
    \mathcal{L}_{\mathrm{LMM}}^{\mathrm{text}} (\Theta = \{\theta, \xi, \phi\}, \bm{x}, \bm{I}) = \frac{-1}{T-N}\sum_{i=N+1}^T \log p_{\Theta}(\bm{x}_i | \bm{x}_{<i}, \bm{v}),
\end{equation}
where $N$ represents the number of visual tokens and visual outputs (one input token corresponds to one visual output).
From \Cref{eq:lmm}, we can tell that only text outputs $\bm{x}_{i>N}$ are supervised.

\section{\method: Reconstructive Visual Instruction Tuning}
\vspace{-5pt}
In this section, we first provide an overview of our reconstructive visual instruction tuning (\method).
Then, we discuss our explorations towards the optimal formulation in the following subsections, with the ultimate goal of \textit{handling spatial redundancy of visual signals} to provide meaningful visual supervision.
Our explorations mainly include reconstruction \textit{targets} and the training \textit{objective}.
\begin{figure}[H]
\vspace{-8pt}
\begin{minipage}{0.42\linewidth}
    \myparagraph{Overview.}
    Illustrated in \Cref{fig:overview}, the overall philosophy of our \method is to construct \textit{reconstructive} visual supervision signals on visual outputs $\bm{x}_{i\leq N}$.
    The training objective includes \textbf{\textit{(i)}} the original next-token prediction on $\bm{x}_{i>N}$ shown in the right part of \Cref{fig:overview}, and \textbf{\textit{(ii)}} another \textit{reconstructive term} in the \textit{left} part of \Cref{fig:overview}, \textit{i.e.}, $\mathcal{L}_{\mathrm{Ross}} = \mathcal{L}_{\mathrm{LMM}}^{\mathrm{text}} + \mathcal{L}_{\mathrm{LMM}}^{\mathrm{visual}}$.
    Specifically, this visual term could be any custom measurements $\mathcal{M}$ between $\bm{x}_{i\leq N}$ and specific reconstruction targets of image $\bm{I}$:
    \begin{equation}
    \label{eq:method}
    \begin{aligned}
    \mathcal{L}_{\mathrm{LMM}}^{\mathrm{visual}} &(\Theta = \{\theta, \xi, \phi, \pi\}, \bm{x}, \bm{I}) \\
    &= \mathcal{M}(\mathcal{J}_{\pi}(\bm{x}_{i\leq N}), \mathcal{F}(\bm{I})),
    \end{aligned}
    \end{equation}
    where $\mathcal{J}_{\pi}$ indicates the $\pi$-parameterized post projection that maps the dimensions of visual tokens $\bm{x}_{i\leq N}$ to be consistent with the teacher tokenizer $\mathcal{F}$. 
\end{minipage}
\hfill
\begin{minipage}{0.55\linewidth}
    \centering
    \includegraphics[width=1\linewidth]{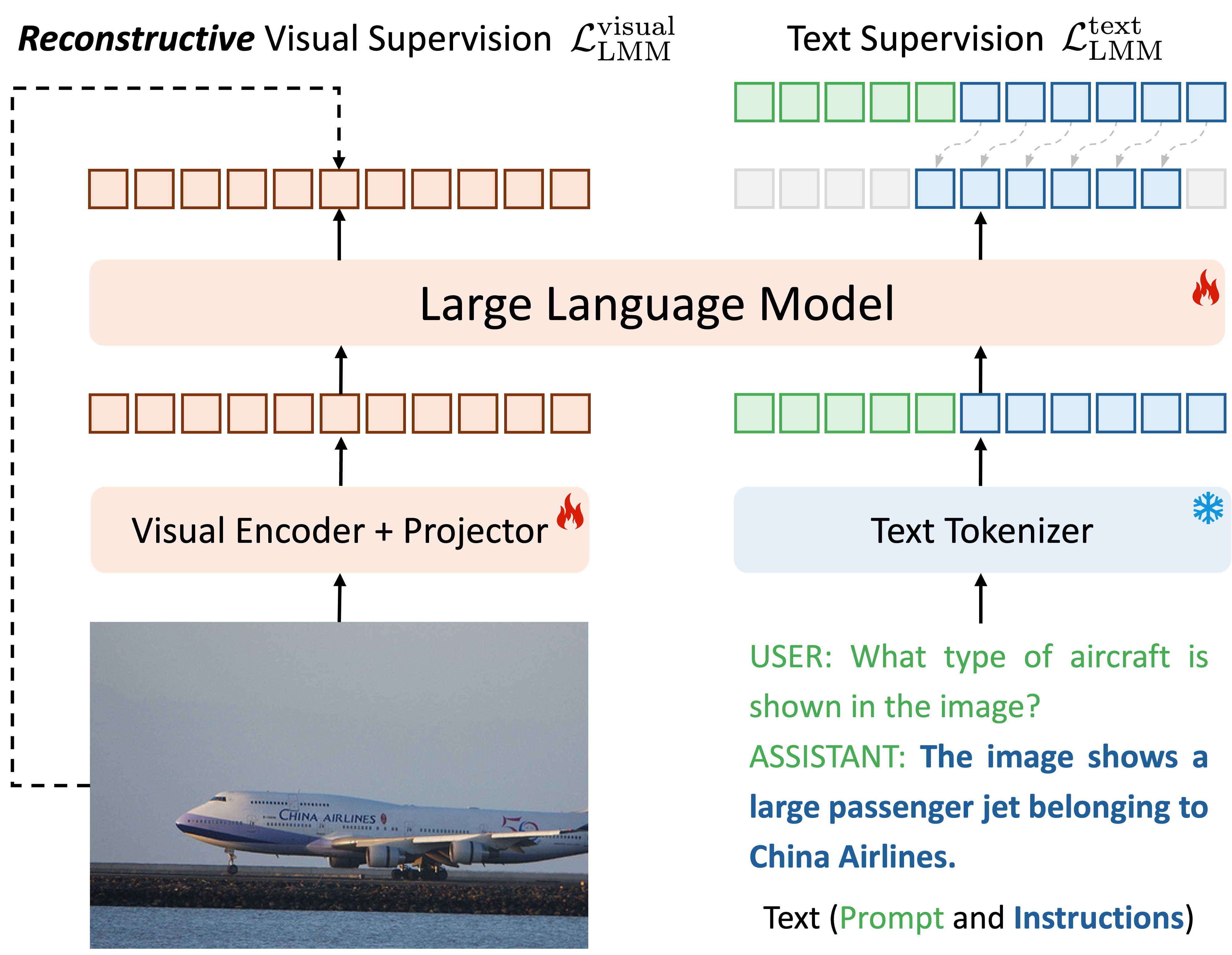}
    \vspace{-15pt}
    \caption{
    \textbf{Overview of \method.}
    Given an input image and the corresponding text to this image, \method aims to \textit{supervise visual outputs by reconstruction.}
    %
    }
    \label{fig:overview}
\end{minipage}
\vspace{-10pt}
\end{figure}

\myparagraph{Variants of \method.}
Evidently, different choices of $\mathcal{F}$ and $\mathcal{M}$ contribute to different variants.
$\mathcal{F}$ controls the reconstruction \textit{target} while $\mathcal{M}$ defines the \textit{objective}:
\begin{enumerate}
    \item Towards the \textit{target}, $\mathcal{F}$ can be the pachify operation~\citep{dosovitskiy2021image}, resulting in \textit{pixel-level} reconstruction, or pre-trained fine-grained visual tokenizers such as VAE~\citep{kingma2013auto} and VQGAN~\citep{esser2021taming}, leading to \textit{latent-level} reconstruction.
    $\mathcal{F}$ could even be vision-only models such as DINOv2~\citep{oquab2023dinov2}, making LMMs learn specific visual patterns from $\mathcal{F}$, which is also a type of \textit{latent-level} reconstruction.
    \item Towards the \textit{objective}, the most straightforward choice of $\mathcal{M}$ is MSE or cosine similarity for \textit{regressing} raw pixel values or latent features, respectively.
    We also explore the \textit{denoising} objective~\citep{ho2020denoising} to avoid being overwhelmed by fitting exact values.
\end{enumerate}
We introduce our explorations step by step in the following sections.
The ultimate goal of our exploration is to design an appropriate self-supervised reconstructive pre-text task that provides meaningful vision-centric supervision signals to LMMs, where handling the \textit{spatial redundancy} of visual signals~\citep{he2022masked} becomes the crux.

\subsection{\textbf{Ross}$^{\text{R}}$: Regressing as Reconstructive Visual Instruction}
\begin{figure}[t]
\centering
\begin{minipage}{0.32\linewidth}
    \includegraphics[width=1\linewidth]{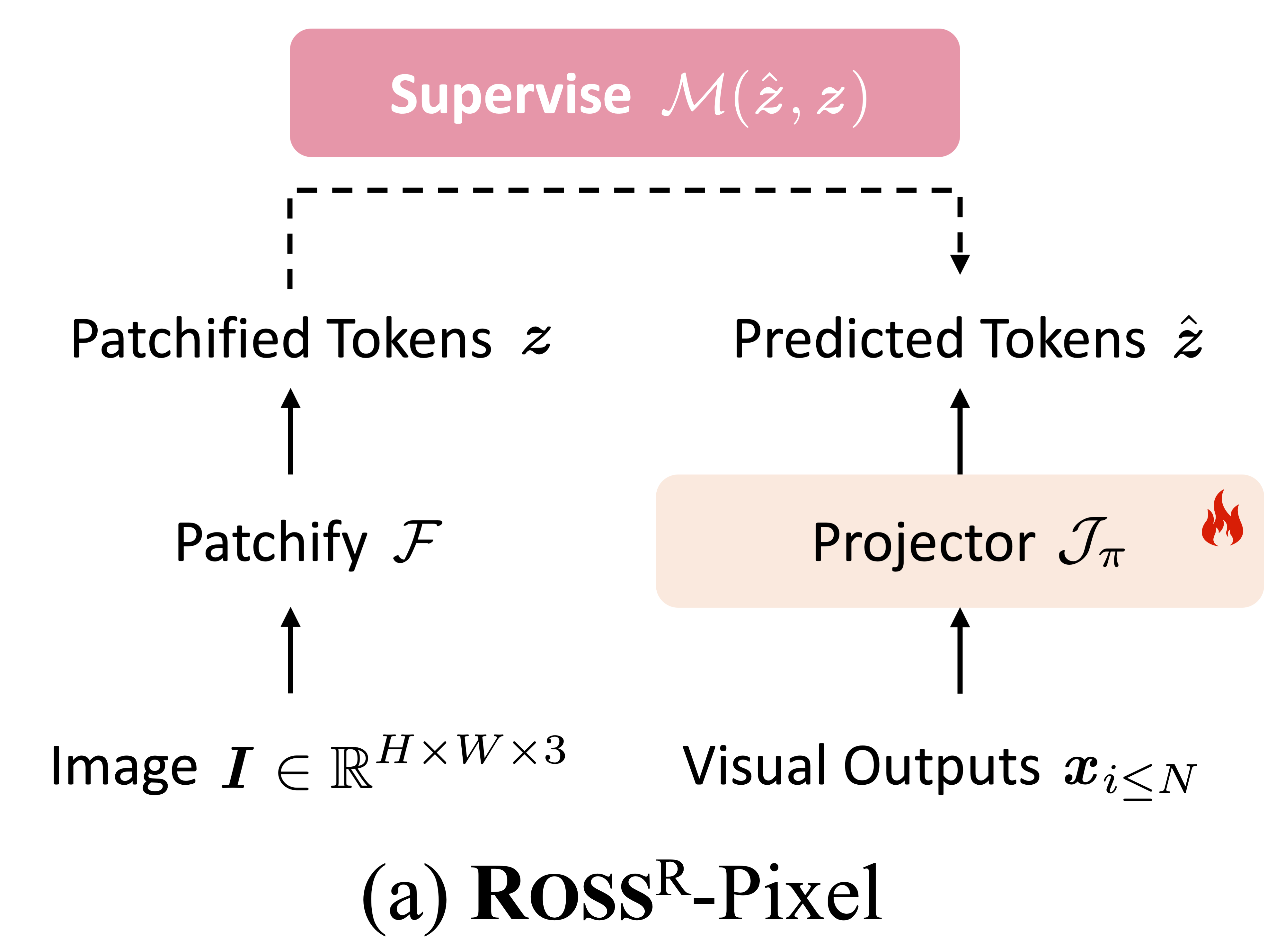}
\end{minipage}
\hfill
\begin{minipage}{0.32\linewidth}
    \includegraphics[width=1\linewidth]{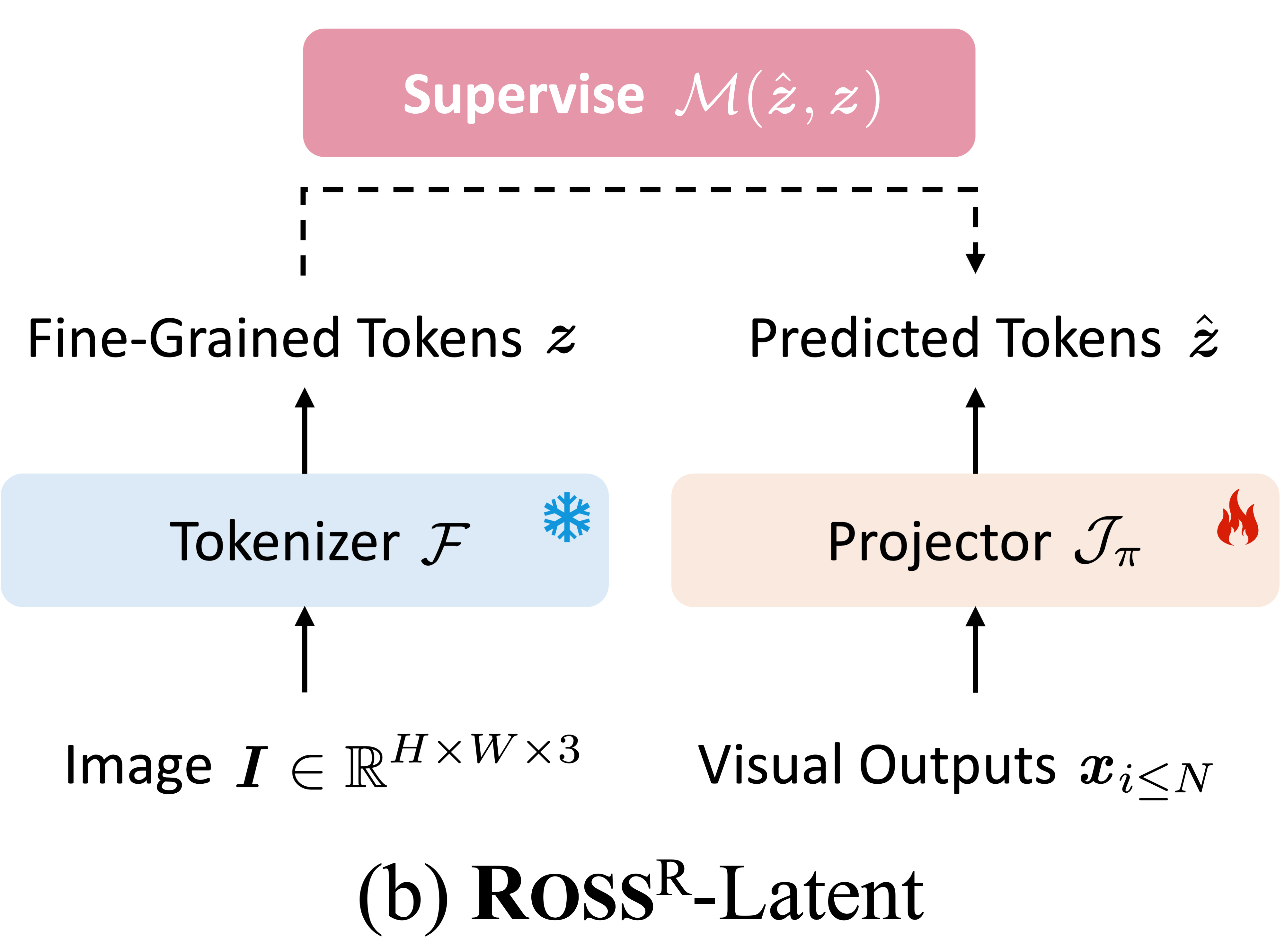}
\end{minipage}
\hfill
\begin{minipage}{0.32\linewidth}
    \includegraphics[width=1\linewidth]{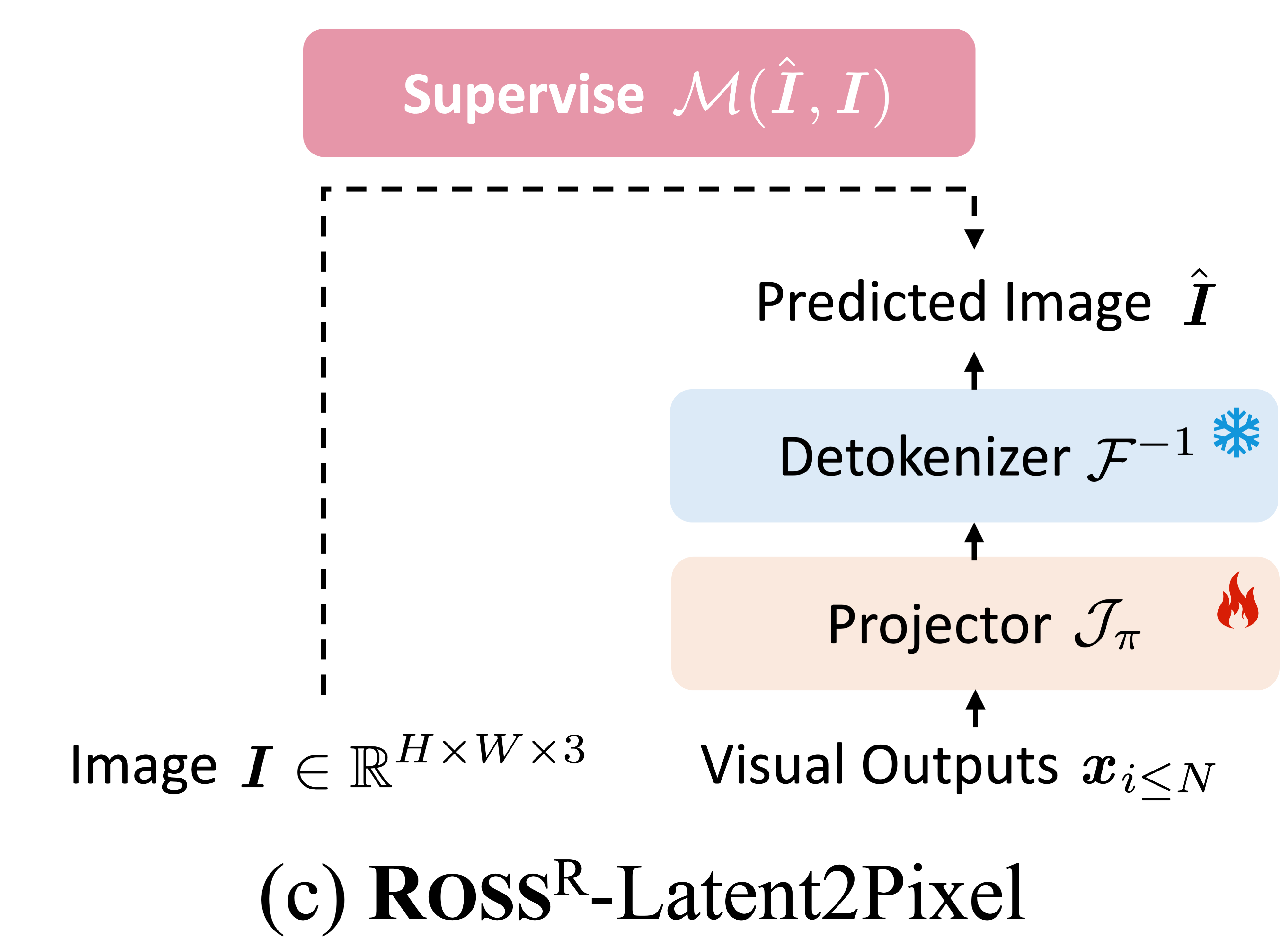}
\end{minipage}
\vspace{-7pt}
\caption{
Variants of {\sc\textbf{Ross}}$^{\text{R}}$, where \textit{regression} objectives are either computed on raw RGB values in (a) and (c), or specific latent space determined by $\mathcal{F}$ in (b).
We adopt MSE as $\mathcal{M}$ for \textit{pixel} regression in (a) and (c), and cosine-similarity for \textit{latent} regression in (b), respectively.
}
\label{fig:RossR}
\vspace{-10pt}
\end{figure}

\vspace{-5pt}
In this section, we introduce straightforward variants, \textit{i.e.},  \textit{regressing} as reconstructive visual instruction.
As shown in \Cref{fig:RossR}, depending on the choice of $\mathcal{F}$, it mainly has three variants: (a) {\sc\textbf{Ross}}$^{\text{R}}$-Pixel, (b) {\sc\textbf{Ross}}$^{\text{R}}$-Latent, and (c) {\sc\textbf{Ross}}$^{\text{R}}$-Latent2Pixel.

\myparagraph{Directly Regressing Raw RGB Values.}
The most straightforward variant is to directly regress raw RGB values illustrated in \Cref{fig:RossR}\textcolor{red}{a}, called ``{\sc\textbf{Ross}}$^{\text{R}}$-Pixel''.
Under such a setting, $\mathcal{F}$ is the patchify operation~\citep{dosovitskiy2021image}, reshaping the image $\bm{I} \in \mathbb{R}^{H \times W \times 3}$ into a sequence of flattened 2D patches $\bm{I}_p \in \mathbb{R}^{N \times (3P^2)}$, where $(P, P)$ is the resolution of each image patch and $N=HW/P^2$ indicates the resulting number of patches.
$\mathcal{J}_{\pi}$ can be a simple MLP, mapping the dimension of visual outputs $\bm{x}_{i \leq N}$ from $D$ to $3P^2$. 
The measurement $\mathcal{M}$ is MSE.
However, as visual signals suffer from \textit{heavy spatial redundancy}~\citep{he2022masked}, such a design may not provide meaningful supervision to LMMs.
An intuitive alternative to avoid directly regressing raw RGB values while still reconstructing the image is to urge LMMs to reconstruct \textit{latent tokens}, introduced as follows.

\myparagraph{Regressing Latent Tokens.}
Illustrated in \Cref{fig:RossR}\textcolor{red}{b}, {\sc\textbf{Ross}}$^{\text{R}}$-Latent aims to regress fine-grained \textit{latent} tokens extracted by the teacher tokenizer.
$\mathcal{F}$ can be models trained with discriminative tasks such as DINOv2~\citep{oquab2023dinov2} and DEIT-III~\citep{touvron2022deit}.
The \textit{encoder} part of models trained with reconstruction tasks such as VQGAN~\citep{esser2021taming} and VAE~\citep{kingma2013auto} are also capable.
$\mathcal{M}$ here is the consine-similarity.
Intuitively, the \textit{decoder} part of the latter is able to remap latent tokens into the pixel space.
Therefore, supervising in the pixel space via \textit{decoding} becomes another valid variant introduced as follows.

\myparagraph{Regressing RGB Values via Decoding.}
Shown in \Cref{fig:RossR}\textcolor{red}{c}, {\sc\textbf{Ross}}$^{\text{R}}$-Latent2Pixel requires a \textit{decoder} to project predicted latent tokens $\hat{\bm{z}}$ into the RGB pixel space, resulting in predicted image $\hat{\bm{I}}$.
Let $\mathcal{F}^{-1}$ be the \textit{decoder} part of VQGAN~\citep{esser2021taming} or VAE~\citep{kingma2013auto}, and the \textit{regressive} MSE objective $\mathcal{M}$ is performed on pixel-space.
Note that we simply use $\mathcal{F}^{-1}$ to represent the decoding process, which is actually \textit{not} the inverse function of $\mathcal{F}$ mathematically.

\myparagraph{Discussion.}
Recall that we need to find the optimal solution to address the spatial redundancy of natural visual signals, the \textit{target-level} exploration above achieves this goal \textit{partially}, as the \textit{objective} is limited to vanilla regression.
To this end, inspired by~\cite{ho2020denoising} and~\cite{li2024autoregressive}, we further incorporate a novel \textit{denoising} objective in the following section.

\subsection{\textbf{Ross}$^{\text{D}}$: Denoising as Reconstructive Visual Instruction}
\vspace{-5pt}
As an objective for handling \textit{heavy spatial redundancy} to provide meaningful vision-centric supervision signals, denoising is better than vanilla regressing, since the introduction of noise into the training data acts as an implicit form of data augmentation and regularization.
The denoising process encourages the model to focus on the underlying data manifold rather than memorizing specific instance values~\citep{chen2023score, song2019generative, karras2022elucidating, yang2024the}.
%

\begin{figure}
    \centering
    \includegraphics[width=1\linewidth]{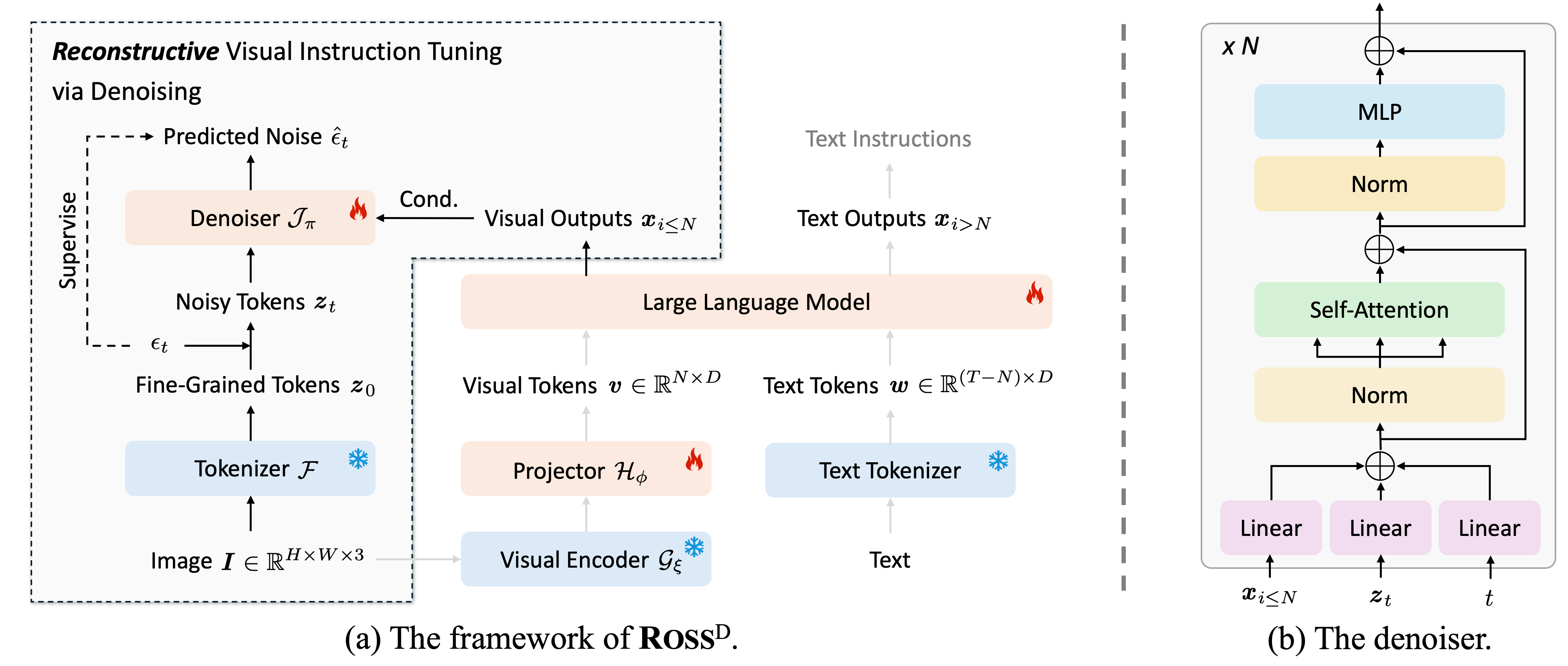}
    \vspace{-15pt}
    \caption{
    Illustration of (a) the training procedure of {\sc\textbf{Ross}}$^{\text{D}}$ and (b) the detailed architecture of the denoiser $\mathcal{J}_{\pi}$. 
    (a) {\sc\textbf{Ross}}$^{\text{D}}$ introduces visual guidance via \textit{denoising fine-grained visual tokens $\bm{z}_0$ conditioning on visual outputs $\bm{x}_{i \leq N}$}.
    (b) The denoiser takes noisy tokens $\bm{z}_t$, current timesteps $t$, and conditions $\bm{x}_{i \leq N}$ as inputs and outputs the predicted noise $\hat{\epsilon}_t$.
    Each denoiser block consists of three linear projection layers and a standard self-attention block~\citep{vaswani2017attention}.
    %
    }
    \label{fig:method}
    \vspace{-10pt}
\end{figure}

Techinically, as illustrated in \Cref{fig:method}\textcolor{red}{a},
our final {\sc\textbf{Ross}}$^{\text{D}}$ takes high-level visual outputs $\bm{x}_{i \leq N}$ as conditions to recover \textit{clean} fine-grained tokens $\bm{z}_0$ from \textit{noisy} tokens $\bm{z}_t$.
Specifically, clean tokens $\bm{z}_0 = \mathcal{F}(\bm{I})$ are obtained from the teacher tokenizer $\mathcal{F}$.
By default, we utilize a continuous VAE~\citep{kingma2013auto} regularized by Kullback–Leibler (KL) divergence provided by~\cite{rombach2022high}, since it is believed to capture sufficient image details.
The training procedure of the denoiser $\mathcal{J}_{\pi}$ follows a diffusion process~\citep{ho2020denoising}:
\begin{equation}
    \label{eq:visual}
    \mathcal{L}_{\mathrm{LMM}}^{\mathrm{visual}} (\Theta = \{\theta, \xi, \phi, \pi\}, \bm{x}, \bm{I}) = \mathbb{E}_{t, \bm{\epsilon}} \left[
    || \mathcal{J}_{\pi}(\bm{z}_t; \mathbf{x}_{i \leq N}, t) - \bm{\epsilon} ||^2
    \right].
\end{equation}
The denoiser $\mathcal{J}_{\pi}$ actually estimates the conditional expectation $\mathbb{E}[\bm{\epsilon} \sim \mathcal{N}(\bm{0}, \mathbf{I}) | \bm{z}_t]$.
More details about the background knowledge of diffusion models can be found in \Cref{sec:more_pre}.

\myparagraph{Architecture of the Denoiser.}
As conditions $\bm{x}_{i \leq N}$ are \textit{causal}, we introduce a self-attention module to model the inter-token dependencies illustrated in \Cref{fig:method}\textcolor{red}{b}.
Specifically, the architecture of the denoiser $\mathcal{J}_{\pi}$ is a stack of Transformer Encoder blocks~\citep{vaswani2017attention} and each block contains three extra projections for conditions $\bm{x}_{i \leq N}$, inputs $\bm{z}_t$, and timesteps $t$, respectively.
%

\myparagraph{Choices of the Teacher Tokenizer.}
By default, we adopt \textit{latent} denoising and we take a continuous tokenizer provided by~\cite{rombach2022high} as $\mathcal{F}$, since it manages to reconstruct input images with a low rFID~\citep{heusel2017gans} and thus it is expected to preserve many low-level details of input images.
This extra reconstructive objective, however, is \textit{not} limited to any certain tokenizer $\mathcal{F}$.
Discrete tokenizers such as VQGAN~\citep{esser2021taming}, and vision self-supervised models such as DINOv2~\citep{oquab2023dinov2}, are also qualified to be the tokenizer.
Even the patchify operation~\citep{dosovitskiy2021image} is capable, resulting in \textit{pixel} denoising.

\section{Experiments}
\vspace{-5pt}
%

\subsection{Ablation Study}
\label{sec:ablation}
\vspace{-5pt}
\myparagraph{Implementation Details.}
All ablation studies are implemented based on LLaVA-v1.5~\citep{liu2024improved}.
The visual encoder $\mathcal{G}_{\xi}$ is CLIP-ViT-L/14@336~\citep{radford2021learning} and the base LLM is Qwen2-7B-Instruct~\citep{yang2024qwen2}.
The training data is LLaVA-558K~\citep{liu2023visual} and Cambrian-737K~\citep{tong2024cambrian} for the pre-training stage and the instruction tuning stage, respectively.
%
%
We evaluate our each variant of \method mainly on \textit{\textbf{(i)}} hallucination: POPE~\citep{li2023evaluating} and HallusionBench~\citep{guan2024hallusionbench}, \textbf{\textit{(ii)}} fine-grained comprehension: MMVP~\citep{tong2024eyes} and ChartQA~\citep{masry2022chartqa}, and \textbf{\textit{(iii)}} general comprehension:  MMBench~\citep{liu2023mmbench} English dev split.
All evaluations are conducted with VLMEvalKit~\citep{duan2024vlmevalkit}.
Evaluation prompts can be found in \Cref{sec:impl_details}.

\begin{figure}[t]
    \centering
    
    \includegraphics[width=1\linewidth]{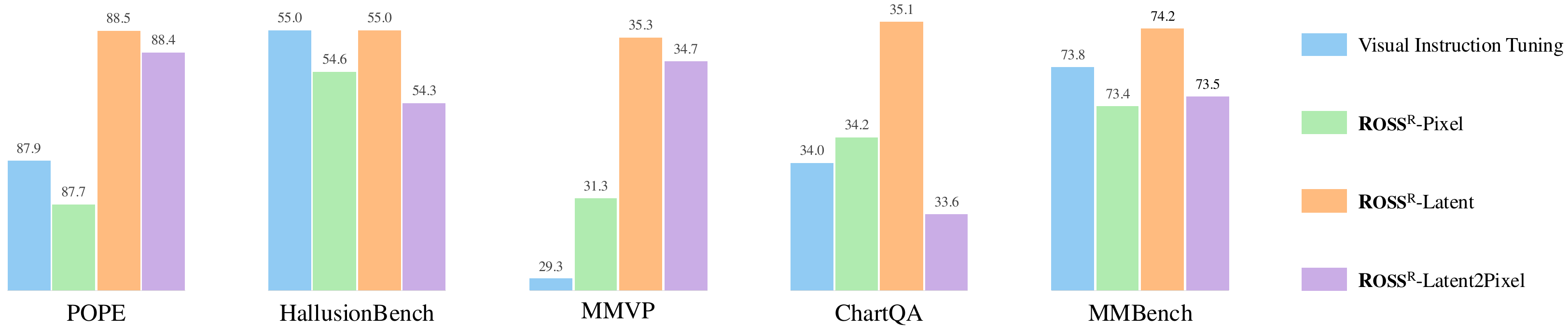}
    \vspace{-18pt}
    \caption{
    \textbf{Pixel Regression \textit{v.s.} Latent Regression.}
    The teacher tokenizer $\mathcal{F}$ for {\sc\textbf{Ross}}$^{\text{R}}$-Latent is the \textit{encoder} of a continuous VAE~\citep{kingma2013auto} provided by~\cite{rombach2022high}, while its \textit{decoder} serves as $\mathcal{F}^{-1}$ for {\sc\textbf{Ross}}$^{\text{R}}$-Latent2Pixel.
    Our vision-centric reconstructive supervision surpasses the visual instruction tuning baseline in most cases.
    Among three regression variants, {\sc\textbf{Ross}}$^{\text{R}}$-Latent performs the best, as it avoids explicitly regressing redundant raw RGB values.
    }
    \label{fig:pixel_latent}

    \vspace{5pt}

    \includegraphics[width=1\linewidth]{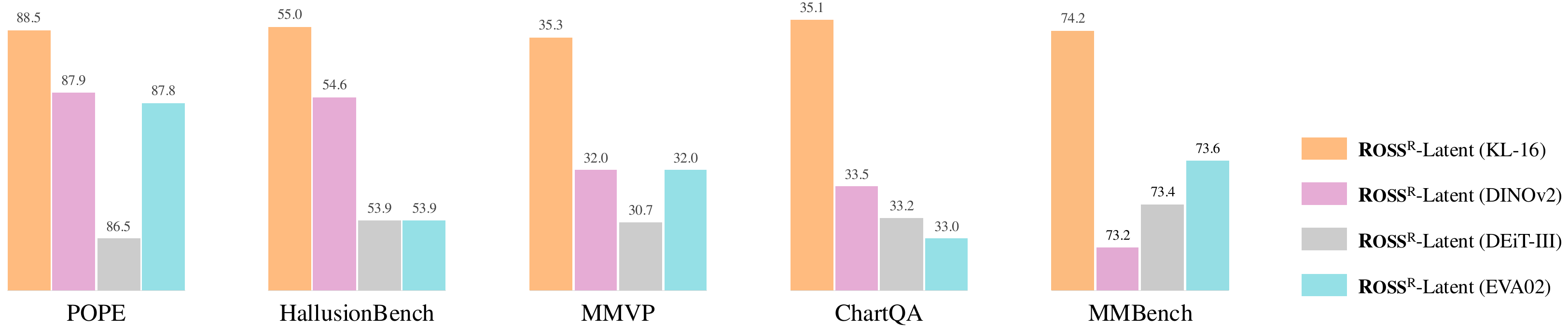}
    \vspace{-18pt}
    \caption{
    \textbf{Choices of the latent teacher tokenizer $\mathcal{F}$.}
    KL-16~\citep{rombach2022high} is the best tokenizer as it is originally used for \textit{reconstruction}, and it is expected to preserve the most image details.
    Other alternatives are utilized for classification~\citep{touvron2022deit}, instance-level representation learning~\citep{oquab2023dinov2}, and language alignment~\citep{fang2024eva}, respectively.
    }
    \label{fig:reg_tokenizer}

    \vspace{5pt}

    \includegraphics[width=1\linewidth]{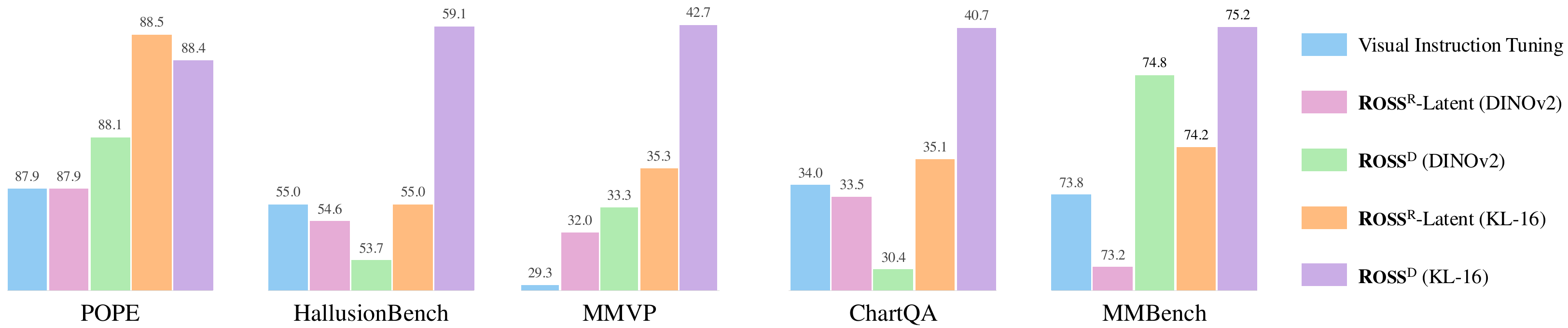}
    \vspace{-18pt}
    \caption{
    \textbf{Regression \textit{v.s.} Denoising.}
    With KL-16 as the tokenizer, the denoising objective introduced in \Cref{eq:visual} brings significant improvements over vanilla regression using MSE as it avoids overfitting exact latent token values, even if {\sc\textbf{Ross}}$^{\text{R}}$-Latent (KL-16) has already outperformed the visual instruction tuning baseline by a large margin.
    }
    \label{fig:denoising_regression}

    \vspace{10pt}
    
    \includegraphics[width=1\linewidth]{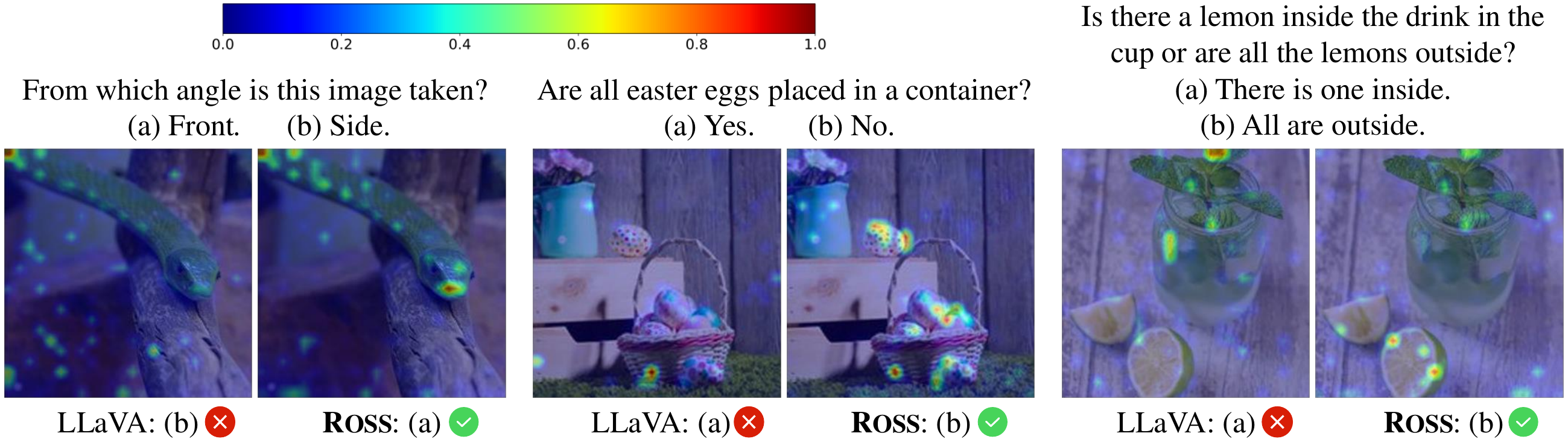}
    \vspace{-15pt}
    \caption{
    \textbf{Qualitative comparison on attention maps} on MMVP~\citep{tong2024eyes}, where we keep the \textit{same} LLM and training data.
    With extra vision-centric supervision signals, \method urges the model to \textit{focus on specific image contents corresponding to the question} with higher attention values.
    }
    \label{fig:attention}

    \vspace{-10pt}
\end{figure}

\myparagraph{Pixel Regression \textit{v.s.} Latent Regression.}
Starting from the visual instruction tuning baseline~\citep{liu2023visual, liu2024improved}, we first explore the effectiveness of using \textit{regression} as the objective for our reconstructive visual instruction tuning.
%
%
We utilize a continuous VAE~\citep{kingma2013auto} with an encoder-decoder architecture provided by~\cite{rombach2022high}, where the \textit{encoder} part serves as $\mathcal{F}$ for {\sc\textbf{Ross}}$^{\text{R}}$-Latent while the \textit{decoder} part is $\mathcal{F}^{-1}$ for {\sc\textbf{Ross}}$^{\text{R}}$-Latent2Pixel.
As illustrated in \Cref{fig:pixel_latent}, our vision-centric regression supervision outperforms the visual instruction tuning baseline in most cases.
Moreover, latent regression performs the best since \textit{regressing raw RGB pixels fails to provide meaningful supervision signals}, regardless of whether utilizing a decoder or not.

\myparagraph{Choices of $\mathcal{F}$.}
We study the effectiveness across different latent teacher tokenizers $\mathcal{F}$ in \Cref{fig:reg_tokenizer},  including KL-16 provided by~\cite{rombach2022high}, which is a continuous VAE~\citep{kingma2013auto} with Kullback–Leibler (KL) divergence, self-supervised DINOv2~\citep{oquab2023dinov2}, fully-supervised DEiT-III~\citep{touvron2022deit}, and language-supervised EVA02CLIP~\citep{fang2024eva}.
Among them, KL-16 is the best choice.
One intuitive explanation is that it is expected to preserve the most image details, since it was originally designed to accurately reconstruct input images.

\myparagraph{Regression \textit{v.s.} Denoising.}
In \Cref{fig:denoising_regression}, we study the effectiveness of the \textit{denoising} objective over vanilla regression across different tokenizers, \textit{i.e.}, KL-16~\citep{rombach2022high} and DINOv2~\citep{oquab2023dinov2}.
Notably, even if {\sc\textbf{Ross}}$^{\text{R}}$-Latent (KL-16) has already outperformed the visual instruction tuning baseline by a large margin, {\sc\textbf{Ross}}$^{\text{D}}$ manages to bring significant improvements by replacing regression with denoising.
Therefore, \textit{denoising is better at handling visual spatial redundancy}.

Finally, we leverage the insights and conclusions from all our previous studies to train our \method.
Specifically, we regard the optimal formulation {\sc\textbf{Ross}}$^{\text{D}}$ (KL-16), \textit{i.e.}, denoising with the KL-16 tokenizer, as our final \method.
\textcolor{textpurple}{
Please refer to \Cref{sec:more_abl} for ablations on the architecture of the denoiser, continuous tokenizer \textit{v.s.} discrete tokenizer, and the denoising schedule.
}

\subsection{In-Depth Analysis}
\label{sec:depth}
\vspace{-5pt}
\begin{table}[H]
\vspace{-5pt}
\begin{minipage}{0.53\linewidth}
\myparagraph{Attention Analysis.}
We compute the attention scores of the \textit{last} token with respect to \textit{all visual tokens} on MMVP~\citep{tong2024eyes}.
Quantitative and qualitative comparisons between the visual instruction tuning baseline (LLaVA)~\citep{liu2024improved} and our \method are provided in \Cref{tab:attention} and \Cref{fig:attention}, respectively.
\Cref{tab:attention} reveals that the attention scores achieved by \method are \textit{significantly higher} than those of LLaVA, indicating that the inclusion of vision-centric reconstructive objective $\mathcal{L}_{\mathrm{LMM}}^{\mathrm{visual}}$ effectively directs focus towards input images, thereby enhancing the comprehending visual signals.
Similarly, \Cref{fig:attention} demonstrate that the implementation of $\mathcal{L}_{\mathrm{LMM}}^{\mathrm{visual}}$ enables the alignment of attention closely with the \textit{relevant visual elements} corresponding to the text query.
\end{minipage}
\hfill
\begin{minipage}{0.44\linewidth}
    \vspace{-8pt}
    \centering\small
    \caption{
    \textbf{Quantitative comparison on attention values.}
    We conduct a T-test~\citep{student1908probable} to compare the \textit{means} and a Mann-Whitney U test \citep{mann1947test} to compare the \textit{medians} of the two distributions.
    The mean and median of \method are both \textit{significantly higher} than those of LLaVA.
    }
    \vspace{-5pt}
    \label{tab:attention}
    \setlength{\tabcolsep}{3pt}
    \begin{tabular}{cccc}
    \toprule
    Statistic ($\times$10$^{-\text{4}}$) & LLaVA & \method & P-value \\
    \midrule
    Mean & 2.03 & 2.36 & 1.27$\times$10$^{-\text{7}}$ \\
    25th Percentile & 1.50 & 1.81 & -- \\
    Median & 1.90 & 2.26 & 4.39$\times$10$^{-\text{9}}$ \\
    75th Percentile & 2.42 & 2.76 & -- \\
    95th Percentile & 3.51 & 3.69 & -- \\
    \bottomrule
    \end{tabular}
\end{minipage}
\vspace{-10pt}
\end{table}

\begin{table}[t]
    \centering\small
    \caption{
    \textbf{Generative \textit{v.s.} Reconstructive.}
    Following~\cite{sun2024generative} and~\cite{dong2024dreamllm}, we adopt 576 learnable latent tokens to \textit{query} the LMM and utilize the corresponding outputs as conditions to the denoiser for generative cases.
    Extra 102K caption data from ShareGPT4V~\citep{chen2023sharegpt4v} is introduced to the original SFT data, facilitating text-to-image creation.
    Reconstructive objectives boost comprehension while generative alternatives \textit{cannot}.
    }
    \label{tab:recon_gen}
    \vspace{-5pt}
    \setlength{\tabcolsep}{4.8pt}
    \begin{tabular}{ll cc cccccc}
    \toprule
    \multirow{2}{*}{Method} & \multirow{2}{*}{SFT Data} & \multicolumn{2}{c}{w/ $\mathcal{L}_{\mathrm{LMM}}^{\mathrm{visual}}$} & \multicolumn{2}{c}{Hallucination} & \multicolumn{2}{c}{Fine-Grained} & General \\
    \cmidrule(lr){3-4} \cmidrule(lr){5-6} \cmidrule(lr){7-8} \cmidrule(lr){9-9}
    & & 737K & 102K & POPE & Hallu. & MMVP & ChartQA & MMB$^{\text{EN}}$ \\
    \midrule
    Baseline & 737K + \textit{Caption} 102K & -- & -- & 86.2 & 55.1 & 32.0 & 30.9 & 74.4 \\
    Reconstructive & 737K + \textit{Caption} 102K & \checkmark & \checkmark & \textbf{87.6} & \textbf{58.0} & \textbf{38.7} & \textbf{40.4} & \textbf{75.2} \\
    Reconstructive & 737K + \textit{Caption} 102K & -- & \checkmark & \textbf{87.6} & 56.3 & 37.3 & 39.7 & 74.9 \\
    Generative & 737K + \textit{Creation} 102K & -- & \checkmark & 85.4 & 52.0 & 30.0 & 31.2 & 73.9 \\
    \bottomrule
    \end{tabular}

    \vspace{15pt}
    
    \caption{
    \textbf{The effectiveness of the vision-centric supervision among various LLMs and visual encoders}, where $\mathcal{L}_{\mathrm{LMM}}^{\mathrm{visual}}$ manages to bring significant improvements \textit{consistently}.
    %
    %
    }
    \label{tab:llm}
    \vspace{-5pt}
    \setlength{\tabcolsep}{5pt}
    \begin{tabular}{rc llllll}
    \toprule
    Language Model & $\mathcal{L}_{\mathrm{LMM}}^{\mathrm{visual}}$ & POPE & Hallu. & MMVP & ChartQA & OCRBench & MMB$^{\text{EN}}$ \\
    \hline
    \rowcolor{lightgray}
    \multicolumn{8}{l}{\textit{Visual Encoder: CLIP-ViT-L/14@336}} \\
    \multirow{2}{*}{Vicuna-7B-v1.5} & -- & 86.3 & 52.5 & 28.0 & 32.9 & 339 & 67.0 \\
    & \checkmark & \textbf{87.2 \up{0.9}} & \textbf{55.8 \up{3.3}} & \textbf{36.0 \up{8.0}} & \textbf{39.8 \up{6.9}} & \textbf{350 \up{11}} & \textbf{67.6 \up{0.6}} \\
    
    \multirow{2}{*}{Qwen2-7B-Instruct} & -- & 87.9 & 55.0 & 29.3 & 34.0 & 363 & 73.8 \\
    & \checkmark & \textbf{88.4 \up{0.5}} & \textbf{56.7 \up{1.7}} & \textbf{42.0 \up{12.7}} & \textbf{37.1 \up{3.1}} & \textbf{381 \up{18}} & \textbf{75.2 \up{1.4}} \\
    
    

    \hline
    \rowcolor{lightgray}
    \multicolumn{8}{l}{\textit{Visual Encoder: SigLIP-ViT-SO400M/14@384}} \\
    \multirow{2}{*}{Vicuna-7B-v1.5} & -- & 86.0 & 50.4 & 27.3 & 36.2 & 354 & 64.5 \\
    & \checkmark & \textbf{86.8 \up{0.8}} & \textbf{53.2 \up{2.8}} & \textbf{38.0 \up{10.7}} & \textbf{41.6 \up{5.4}} & \textbf{365 \up{11}} & \textbf{65.7 \up{1.2}} \\

    \multirow{2}{*}{Qwen2-7B-Instruct} & -- & 88.5 & 57.3 & 40.7 & 44.4 & 432 & 76.3 \\
    & \checkmark & \textbf{88.7 \up{0.2}} & \textbf{58.2 \up{0.9}} & \textbf{49.3 \up{8.6}} & \textbf{46.3 \up{1.9}} & \textbf{448 \up{16}} & \textbf{76.9 \up{0.6}} \\

    
    \bottomrule
    \end{tabular}
    \vspace{-10pt}
\end{table}

\myparagraph{Generative \textit{v.s.} Reconstructive.}
We ablate the effectiveness of reconstruction over generation in \Cref{tab:recon_gen}.
Similar to~\cite{sun2024generative} and~\cite{dong2024dreamllm}, for generative cases, we adopt 576 learnable latent tokens to \textit{query} the LMM and utilize the corresponding outputs as conditions to the denoiser.
The detailed pipeline of these two methods can be found at \Cref{fig:gen_rec} in \Cref{sec:more_abl}.
However, generative methods require \textit{specific creation data} and can \textit{not} be naively implemented on the original SFT data.
To build creation data, we utilize GPT-4o to transfer 102K \textit{caption} into \textit{text-to-image} creation data from ShareGPT4V~\citep{chen2023sharegpt4v} and combine them with the original SFT data.
From \Cref{tab:recon_gen}, we can tell that reconstructive objectives boost comprehension while generative alternatives \textit{cannot}.
\textcolor{textpurple}{
An intuitive explanation of this evidence can be found in \Cref{sec:more_abl}.
}

\myparagraph{\method with Different LLMs and Visual Encoders.}
To demonstrate the effectiveness of our vision-centric supervision $\mathcal{L}_{\mathrm{LMM}}^{\mathrm{visual}}$ adopted by our \method, we conduct systematic experiments across different base LLMs and visual encoders.
%
%
From \Cref{tab:llm}, \method contributes to significant improvements \textit{consistently}, especially on fine-grained comprehension benchmarks, \textit{i.e.}, MMVP~\citep{tong2024eyes} and ChartQA~\citep{masry2022chartqa}.
\textcolor{textpurple}{
Extended experiments on more representative benchmarks can be found at \Cref{tab:extend_abl} in \Cref{sec:more_abl}.
}

\begin{figure}[t]
    \centering
    \includegraphics[width=1\linewidth]{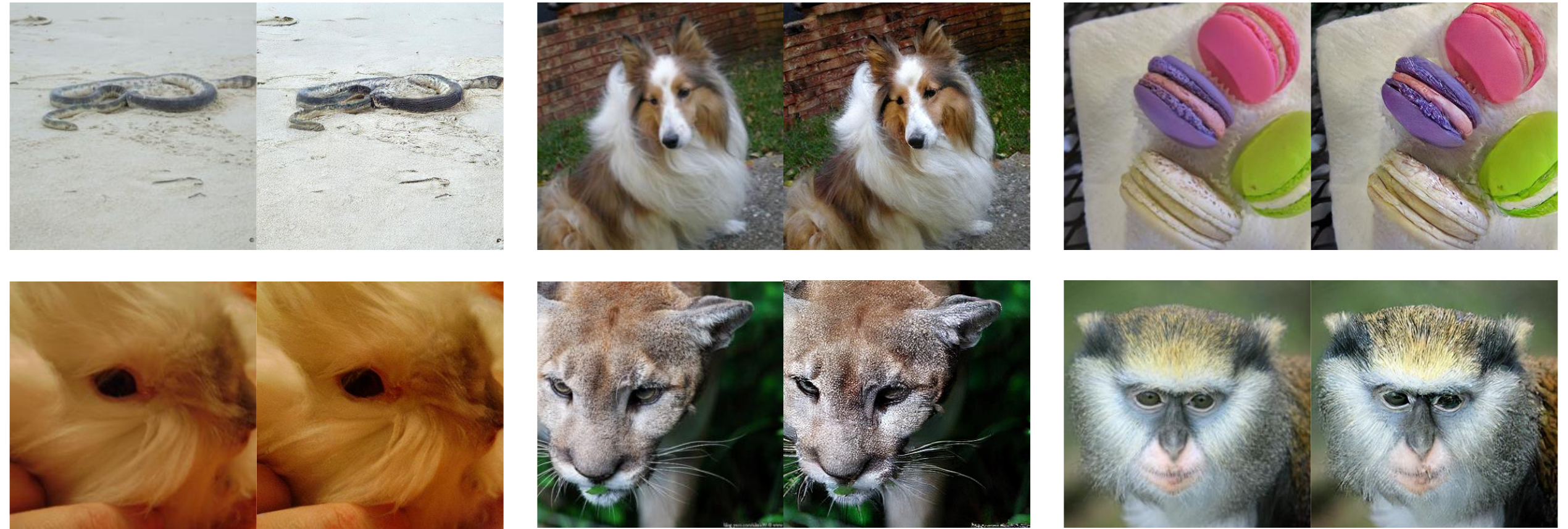}
    \vspace{-15pt}
    \caption{
    \textbf{Reconstruction results} on ImageNet-1K~\citep{deng2009imagenet} validation set.
    For each tuple, we show the input image (left) and the reconstructed image (right).
    %
    %
    Reasonable reconstruction results demonstrate that \textit{high-level features of \textbf{\method-7B} can be projected back into the pixel space.}
    }
    \label{fig:recon}
\end{figure}

\begin{table}[t]
    \centering\small
    \caption{
    \textbf{Comparison to state-of-the-art LMMs}.
    A mixture of 2M caption data and 1.2M instruction tuning data are utilized for pre-training and fine-tuning, respectively.
    Our model outperforms them in most of the settings. 
    We evaluate these models on: POPE \citep{li2023evaluating} averaged accuracy, Hallu.: HallusionBench \citep{guan2024hallusionbench} average accuracy, MMB$^\text{EN}$: MMBench \citep{liu2023mmbench} English dev split, MMB$^\text{CN}$: MMBench \citep{liu2023mmbench} Chinese dev split, SEED$^\text{I}$: SEED-Bench-1 \citep{li2023seed} with image accuracy, MMMU \citep{yue2024mmmu} validation split, MMVP \citep{tong2024eyes}, GQA \citep{hudson2019gqa} test-dev-balanced split, and AI2D~\citep{hiippala2021ai2d} test split.
    $^\ddag$We evaluate the official checkpoint/api using VLMEvalKit~\citep{duan2024vlmevalkit}.
    }
    \vspace{-5pt}
    \setlength{\tabcolsep}{2.5pt}
    \resizebox{\textwidth}{!}{
    \begin{tabular}{r lllllllll}
        \toprule
        Model & POPE & Hallu. & MMB$^\text{EN}$ & MMB$^\text{CN}$ & SEED$^\text{I}$ & MMMU & MMVP & GQA & AI2D \\
        \midrule

        GPT-4V-1106 \citep{gpt4v} & 75.4 & 65.8$^{\ddag}$ & 75.8 & 75.1$^{\ddag}$ & 71.6 & 53.8 & 50.0 & 36.8 & 78.2 \\
        Gemini-1.5 Pro \citep{team2023gemini} & -- & -- & 73.6 & -- & 70.7 & 47.9 & -- & -- & -- \\
        MM-1-8B \citep{mckinzie2024mm1} & 86.6 & -- & 72.3 & -- & 69.9 & 37.0 & -- & 72.6 & -- \\

        \midrule
        
        Mini-Gemini-8B \citep{li2024mini} & -- & -- & 72.7 & -- & 73.2 & 37.3 & 18.7 & 64.5 & 73.5 \\
        DeepSeek-VL-7B \citep{lu2024deepseek} & 85.8$^\ddag$ & 44.1$^\ddag$ & 73.2 & 72.8 & 70.4 & 36.6 & -- & -- & 64.9$^\ddag$ \\
        Cambrian-1-8B \citep{tong2024cambrian} & 87.4$^\ddag$ & 48.7$^\ddag$ & 75.9 & 68.9$^\ddag$ & \textbf{74.7} & 42.7 & 51.3 & 64.6 & 73.0 \\
        
        \textbf{\method-7B} & \textbf{88.3} & \textbf{57.1} & \textbf{79.1} & \textbf{77.1} & 73.6 & \textbf{46.6} & \textbf{56.7} & \textbf{65.5} & \textbf{79.3} \\
            
        \hline
        \rowcolor{lightgray}
        \multicolumn{10}{l}{\textit{Base LLM: Vicuna-7B-v1.5}} \\
        
        LLaVA-v1.5-7B$^\ddag$ \citep{liu2024improved} & 86.2 & 47.5 & 65.5 & 58.5 & 66.0 & 34.4 & 20.0 & 62.0 & 55.4 \\
        LLaVA-v1.6-7B$^\ddag$ \citep{liu2024llavanext} & 86.5 & 35.8 & 67.4 & 60.1 & \textbf{70.2} & 35.8 & 37.3 & \textbf{64.2} & 67.1 \\

        \textbf{\method-7B}$_{\text{vicuna}}$ & \textbf{88.2} & \textbf{55.2} & \textbf{67.7} & \textbf{61.3} & 67.6 & \textbf{36.9} & \textbf{39.3} & 63.7 & \textbf{69.3} \\

        \hline
        \rowcolor{lightgray}
        \multicolumn{10}{l}{\textit{Base LLM: Vicuna-13B-v1.5}} \\
        LLaVA-v1.5-13B$^\ddag$ \citep{liu2024improved} & 82.5 & 44.9 & 68.8 & 63.6 & 68.2 & 36.6 & 32.0 & 63.3 & 60.8 \\
        LLaVA-v1.6-13B$^\ddag$ \citep{liu2024llavanext} & 86.2 & 36.7 & 70.0 & 64.1 & 71.9 & 36.2 & 35.3 & \textbf{65.4} & 72.4 \\
        Mini-Gemini-13B \citep{li2024mini} & -- & -- & 68.6 & -- & 73.2 & 37.3 & 19.3 & 63.7 & 70.1 \\
        Cambrian-1-13B \citep{tong2024cambrian} & 85.7$^\ddag$ & 54.0$^\ddag$ & \textbf{75.7} & 65.9$^\ddag$ & \textbf{74.4} & 40.0 & 41.3 & 64.3 & 73.6 \\
        
        \textbf{\method-13B}$_{\text{vicuna}}$ & \textbf{88.7} & \textbf{56.4} & 73.6 & \textbf{67.4} & 71.1 & \textbf{41.3} & \textbf{44.7} & 65.2 & \textbf{73.8} \\
        \bottomrule
    \end{tabular}
    }
    \label{tab:main}
    \vspace{-10pt}
\end{table}

{\color{textpurple}

\myparagraph{Reconstruction Results.}
We fine-tune the denoiser to recover latent tokens from a frozen KL-16 provided by~\cite{rombach2022high} conditioned on \textbf{\method-7B} features on ImageNet-1K~\citep{deng2009imagenet} for \textit{only five epochs}, where the denoiser manages to produce reasonable reconstruction results as illustrated in \Cref{fig:recon}.
This interesting finding demonstrates that high-level \textbf{\method-7B} features \textit{actually contain image details}.
%
%
We hope this finding will inspire future work.

\myparagraph{Computational Overhead.}
The denoising process introduces a negligible increase in training time ($\approx$10\% compared to the baseline), while the benefits outweigh the minor additional costs.
Please refer to \Cref{tab:cost} in \Cref{sec:impl_details} for details.
}

\subsection{Comparison with State-of-the-Arts}
\label{sec:sota}
\vspace{-5pt}
\method utlizes a \textit{single} SigLIP-ViT-SO400M/14@384~\citep{zhai2023sigmoid} as the visual encoder.
\textbf{\method-7B} utilizes Qwen2-7B-Instruct~\citep{yang2024qwen2} and \textbf{\method-13B}$_{\text{vicuna}}$ adopts Vicuna-13B-v1.5~\citep{vicuna2023} as the base LLM.
The implementation almost follows LLaVA-v1.5~\citep{liu2024improved} \textit{without} the high-resolution image-slicing technique~\citep{liu2024llavanext}.
Thus, our primary comparison of \method with alternative methods focuses on benchmarks that do \textit{not} require exceptionally high-resolution inputs.
We use a mixture of 2M caption data for the pre-training stage, which consists of 1246K from ShareGPT4V~\citep{chen2023sharegpt4v} and 707K from ALLaVA~\citep{chen2024allava}.
The instruction tuning data is a mixture of Cambrian-737K~\citep{tong2024cambrian} and SMR-473K~\citep{zhang2024beyond}.
\textcolor{textpurple}{
We further incorporate our \method with the ``anyres'' technique~\citep{liu2024llavanext} and compare with others on high-resolution benchmarks at \Cref{tab:ocr} in \Cref{sec:ocr}.
}

Illustrated in~\Cref{tab:main}, we compare our \method with both private models~\citep{gpt4v, team2023gemini, mckinzie2024mm1} and open-sourced alternatives~\citep{liu2024improved, liu2024llavanext, tong2024cambrian, li2024mini, lu2024deepseek}.
The previous open-source state-of-the-art Cambrian-1~\citep{tong2024cambrian} leverages \textit{extrinsic assistance} that aggregates CLIP~\citep{radford2021learning}, SigLIP~\citep{zhai2023sigmoid}, DINOv2~\citep{oquab2023dinov2}, and ConvNext~\citep{liu2022convnet}.
On the other hand, our \method stands for \textit{intrinsic activation}.
With only a single SigLIP~\citep{zhai2023sigmoid} model as the visual encoder, our \method surpasses Cambrian-1~\citep{tong2024cambrian}, under most cases, \textit{without} a careful choice of the visual experts and naturally maintains a lightweight inference procedure.
\method is also data-efficient compared with Cambrian-1~\citep{tong2024cambrian}, since it requires 7M instruction tuning data.
Notably, \textbf{\method-7B} even surpasses GPT-4V-1106 and Gemini-1.5 Pro on several benchmarks such as POPE~\citep{li2023evaluating}, MMBench~\citep{liu2023mmbench}, and MMVP~\citep{tong2024eyes}.

\begin{table}[t]
    \centering\small
    \caption{
    \textbf{Transfer learning} on SpatialBench~\citep{cai2024spatialbot}.
    %
    %
    ``RGB'' indicates using only RGB images for testing, while ``RGB + D'' represents taking depth maps as extra inputs.
    %
    %
    The performance of GPT-4o is obtained from~\cite{cai2024spatialbot}.
    \textit{LMMs can better comprehend depth maps with $\mathcal{L}_{\mathrm{LMM}}^{\mathrm{visual}}$.}
    }
    \label{tab:depth}
    \vspace{-5pt}
    \setlength{\tabcolsep}{4pt}
    \resizebox{\textwidth}{!}{
    \begin{tabular}{llcc llllll}
    \toprule
    Method & Test Inputs & $\mathcal{L}_{\mathrm{LMM}}^{\mathrm{visual}}$ & MiDaS & Size & Reaching & Position & Existence  & Counting & Average \\
    \midrule
    \multirow{4}{*}{LLaVA} & RGB & -- & -- & 20.0 & 51.7 & 58.8 & 70.0 & 74.6 & 55.0 \\
    & RGB + D & -- & -- & 21.7 \up{1.7} & 45.0 \down{6.7} & 58.8 \textcolor{mygray}{-- 0.0} & 65.0 \down{5.0} & 77.7 \up{3.1} & 53.6 \down{1.4} \\
    %
    & RGB & -- & \checkmark & 21.7 & 60.0 & 64.7 & 80.0 & 84.1 & 62.1 \\
    & RGB + D & -- & \checkmark & 21.7 \textcolor{mygray}{-- 0.0} & 51.7 \down{8.3} & \textbf{70.6 \up{5.9}} & 65.0 \down{15.0} & \textbf{91.1} \up{7.0} & 60.0 \down{2.1} \\
    \midrule
    \multirow{2}{*}{\method} & RGB & \checkmark & -- & 25.0 & 53.3 & 64.7 & 70.0 & 75.3 & 57.7 \\
    & RGB + D & \checkmark & -- & \textbf{28.3 \up{3.3}} & \textbf{65.0 \up{11.7}} & 67.6 \up{2.9} & \textbf{85.0 \up{15.0}} & 84.6 \textbf{\up{8.7}} & \textbf{66.1 \up{8.4}} \\
    \midrule
    \multirow{2}{*}{\textcolor{mygray}{GPT-4o}} & \textcolor{mygray}{RGB} & -- & -- & \textcolor{mygray}{43.3} & \textcolor{mygray}{51.7} & \textcolor{mygray}{70.6} & \textcolor{mygray}{85.0} & \textcolor{mygray}{84.5} & \textcolor{mygray}{67.0} \\
    & \textcolor{mygray}{RGB + D} & -- & -- & \textcolor{mygray}{40.0 $\downarrow$ 3.3} & \textcolor{mygray}{51.7 -- 0.0} & \textcolor{mygray}{61.8 $\downarrow$ 8.8} & \textcolor{mygray}{90.0 $\uparrow$ 5.0} & \textcolor{mygray}{85.2 $\uparrow$ 0.7} & \textcolor{mygray}{65.7 $\downarrow$ 1.3} \\
    \bottomrule
    \end{tabular}}
    \vspace{-10pt}
\end{table}

\subsection{Applications}
\label{sec:gen}
\vspace{-5pt}
\myparagraph{Transfer Learning on Understanding Depth Maps.}
We further evaluate the transfer learning capability of our \method on SpatialBench~\citep{cai2024spatialbot}, which requires the model to understand \textit{depth maps}.
We compare our \method with the visual instruction tuning baseline, with the \textit{same} training data and model architecture.
Also, we compare the effectiveness of the \textit{extrinsic assistance} solution, \textit{i.e.}, combining a depth expert MiDaS-3.0~\citep{birkl2023midas} to visual instruction tuning, with our \textit{intrinsic activation} solution.
Specifically, the pre-training data is LLaVA-558K~\citep{liu2023visual} and the fine-tuning data is SpatialQA-853K~\citep{cai2024spatialbot}, where each conversation contains the RGB image and the \textit{depth} maps extracted by ZoeDepth~\citep{bhat2023zoedepth}.
The visual encoder is CLIP-ViT-L/14@336~\citep{radford2021learning} and the base LLM is Qwen2-7B-Instruct~\citep{yang2024qwen2}.
As demonstrated in \Cref{tab:depth}, our \method manages to make use of the extra depth map, as consistent and significant improvements are observed when taking ``RGB + D'' inputs for testing.
Extrinsic assistance approaches \textit{cannot} take advantage of extra depth maps when testing. 
Even GPT-4o \textit{cannot} fully understand depth maps.
\textcolor{textpurple}{
Qualitative results can be found at \Cref{fig:spatialbench} in \Cref{sec:visual}.
}

\section{Conclusion}
\vspace{-5pt}
This paper introduces reconstructive visual instruction tuning (\method), leveraging a vision-centric \textit{reconstructive} objective to supervise visual outputs.
To avoid being overwhelmed by heavily redundant raw RGB values, we train a denoiser to recover clean latent visual representations conditioning on visual outputs.
Experimentally, the proposed objective indeed brings enhanced comprehension capabilities and reduced hallucinations.
\method outperforms the state-of-the-art under most cases with only a \textit{single} SigLIP~\citep{zhai2023sigmoid} as the visual encoder.
\textcolor{textpurple}{The in-depth analysis demonstrates that high-level features from \textbf{\method-7B} actually contain sufficient details for low-level image reconstruction. 
This finding reveals the possibility of equipping comprehension LMMs with the ability of \textit{naive generation} without the help of generation experts such as Stable Diffusion~\citep{rombach2022high}.
}

\section*{Acknowledgements}
\vspace{-5pt}
The work was supported by the National Science and Technology Major Project of China (No. 2023ZD0121300), the National Natural Science Foundation of China (No. U21B2042, No. 62320106010), the 2035 Innovation Program of CAS, and the InnoHK program.

\clearpage

{\small
\bibliographystyle{iclr2025_conference}
\bibliography{ref}
}

\newpage
\appendix
\section*{Appendix}

\section{Latent Diffusion Models}\label{sec:more_pre}
\vspace{-5pt}
Given a set of clean latent tokens $\bm{z}_0$ drawn from $p(\bm{z})$, the forward diffusion process is a Markov chain that gradually adds random Gaussian noise to the original sample:
\begin{equation}
    q(\bm{z}_t | \bm{z}_{t-1}) = \mathcal{N}(\sqrt{1 - \beta_t} \bm{z}_{t-1}, \beta_t\mathbf{I}), 
\end{equation}
where $\mathcal{N}(\bm{\mu}, \bm{\Sigma})$ denotes the Gaussian distribution, and $t$ indicates discrete timesteps.
$\beta_t \in (0, 1)$ indicates a pre-defined time-dependent variance schedule.
According to~\cite{ho2020denoising}, to admit sampling $\bm{z}_t$ at an arbitrary timestep $t$ directly from $\bm{z}_0$, this transition can be reformulated as 
\begin{equation}
\begin{aligned}
\label{eq:add_noise}
    &q(\bm{z}_t | \bm{z}_0) = \mathcal{N}(\sqrt{\bar{\alpha}_t} \bm{z}_0, (1 - \bar{\alpha}_t) \mathbf{I}), \\
    &\bm{z}_t = \sqrt{\bar{\alpha}_t} \bm{z}_0 + \sqrt{1 - \bar{\alpha}_t} \bm{\epsilon}, \quad \bm{\epsilon} \sim \mathcal{N}(\bm{0}, \mathbf{I}),
\end{aligned}
\end{equation}
where $\alpha_t = 1 - \beta_t$ and $\bar{\alpha}_t = \prod_{i=1}^t \alpha_t$.
A latent diffusion model learns to \textit{reverse} this progressive noise addition process for latent tokens. 
Specifically, to iteratively generate clean tokens $\bm{z}_0$ from pure noise $\bm{z}_T$ conditioned on $\mathcal{C}$, we need to reverse the forward process by
\begin{equation}
    \bm{z}_{t-1} = \frac{1}{\sqrt{\alpha_t}} \left( \bm{z}_t - \frac{1 - \alpha_t}{\sqrt{1 - \bar{\alpha}_t}} \bm{\epsilon}_{\pi}(\bm{z}_t; \mathcal{C}, t) \right) + \sigma_t \bm{\epsilon},
\end{equation}
where a $\pi$-parameterized neural network $\bm{\epsilon}_{\pi}$ is trained to predict the added noise during the forward process.
$\sigma_t$ indicates the posterior noise variance.
The training objective of $\bm{\epsilon}_{\pi}$ is 
\begin{equation}
    \label{eq:dm}
    \mathcal{L}(\pi, \bm{z}_0) = \mathbb{E}_{t, \bm{\epsilon}} \left[
    || \bm{\epsilon}_{\pi} (\sqrt{\bar{\alpha}_t}\bm{z}_0 + \sqrt{1 - \bar{\alpha}_t} \bm{\epsilon}; \mathcal{C}, t) - \bm{\epsilon} ||^2
    \right].
\end{equation}

\section{Implementation Details}
\label{sec:impl_details}
\vspace{-5pt}

\begin{table}[H]
    \centering\small
    \caption{
    \textbf{Hyperparameters of \method.}
    We obtain most of the configurations from~\cite{liu2024improved}.
    }
    \label{tab:impl}
    \vspace{-5pt}
    \setlength{\tabcolsep}{5pt}
    \begin{tabular}{l cc}
    \toprule
    Config & Stage I & Stage II \\
    \midrule
    Trainable parts & projector + denoiser & projector + LLM + denoiser \\
    Frozen parts & visual encoder + LLM + teacher tokenizer & visual encoder + teacher tokenizer \\
    Global batch size & 256 & 128 \\
    Batch size per GPU & 16 & 4 \\
    Accumulated steps & 2 & 4 \\
    DeepSpeed zero stage & 2 & 3 \\
    Learning rate & 1$\times$10$^{-\text{3}}$ & 2$\times$10$^{-\text{5}}$ \\
    Learning rate schedule & \multicolumn{2}{c}{warmup + cosine decay} \\
    Warmup ratio & \multicolumn{2}{c}{0.03} \\
    Weight decay & \multicolumn{2}{c}{0} \\
    Epoch & \multicolumn{2}{c}{1} \\
    Optimizer & \multicolumn{2}{c}{AdamW} \\
    Precision & \multicolumn{2}{c}{bf16} \\
    \bottomrule
    \end{tabular}
\end{table}

\begin{table}[H]
\begin{minipage}{0.48\linewidth}
    \centering\small
    \caption{
    \textbf{Details of the instruction tuning dataset} provided by~\cite{tong2024cambrian}.
    }
    \vspace{-5pt}
    \label{tab:cambrian}
    \setlength{\tabcolsep}{2.5pt}
    \begin{tabular}{ll}\toprule
    Dataset & \# Samples \\
    \midrule
    LLaVA~\citep{liu2023visual} & 158K \\
    ShareGPT~\citep{sharegpt} & 40K \\
    VQAv2~\citep{goyal2017making} & 83K \\
    GQA~\citep{hudson2019gqa} & 72.1K \\
    OKVQA~\citep{marino2019ok} & 9K \\
    OCRVQA~\citep{mishra2019ocr} & 80K \\
    A-OKVQA~\citep{schwenk2022okvqa} & 50K \\
    TextVQA~\citep{singh2019towards} & 21.9K \\
    RefCOCO~\citep{kazemzadeh2014referitgame} & 30K \\
    VG~\citep{krishna2017visual} & 86.4K \\
    DVQA~\citep{kafle2018dvqa} & 13K \\
    DocVQA~\citep{mathew2021docvqa} & 15K \\
    ChartQA~\citep{masry2022chartqa} & 28.1K \\
    AI2 Diagrams~\citep{kembhavi2016diagram} & 15.5K \\
    \bottomrule
    \end{tabular}
\end{minipage}
\hfill
\begin{minipage}{0.48\linewidth}
    \vspace{-28pt}
    \centering\small
    \caption{
    \textbf{Details of the instruction tuning dataset} provided by~\cite{zhang2024beyond}.
    }
    \vspace{-5pt}
    \label{tab:smr}
    \setlength{\tabcolsep}{2.5pt}
    \begin{tabular}{ll}
    \toprule
    Dataset & \# Samples \\
    \midrule
    ScienccQA~\citep{saikh2022scienceqa} & 9K \\
    TextbookQA~\citep{kembhavi2017you} & 9.5K \\
    AI2 Diagrams~\citep{kembhavi2016diagram} & 12.4K \\
    ChartQA~\citep{masry2022chartqa} & 28.3K \\
    DVQA~\citep{kafle2018dvqa} & 200K \\
    ArxivQA~\citep{li2024multimodal} & 100K \\
    GeoQA3~\citep{chen2021geoqa} & 5K \\ Geometry3K~\citep{lu2021inter} & 2.1K \\
    GeoQA+~\citep{cao2022augmented} & 72.3K \\
    MathVision~\citep{wang2024measuring} & 2.7K \\
    TabMWP~\citep{lu2022dynamic} & 30.7K \\
    \bottomrule
    \end{tabular}
\end{minipage}
\vspace{-10pt}
\end{table}

\begin{table}[t]
    \centering\small
    \caption{
    \textbf{Summary of the evaluation benchmarks.}
    Prompts are mostly borrowed from VLMEvalKit~\citep{duan2024vlmevalkit} and lmms-eval~\citep{lmmseval2024}.
    }
    \label{tab:prompt}
    \vspace{-5pt}
    \setlength{\tabcolsep}{10pt}
    \begin{tabular}{ll}
    \toprule
    Benchmark & Response formatting prompts \\
    \midrule
    POPE \citep{li2023evaluating} & -- \\
    HallusionBench \citep{guan2024hallusionbench} & Answer the question using a single word or phrase. \\ 
    MMBench \citep{liu2023mmbench} & Answer with the option's letter from the given choices directly. \\
    SEED-Bench \citep{li2023seed} & Answer with the option's letter from the given choices directly. \\
    MMMU \citep{yue2024mmmu} & Answer with the option's letter from the given choices directly. \\
    MMVP \citep{tong2024eyes} & Answer with the option's letter from the given choices directly. \\
    AI2D \citep{hiippala2021ai2d} & Answer with the option's letter from the given choices directly. \\
    RealWorldQA \citep{xai2024grok} & Answer with the option's letter from the given choices directly. \\
    GQA \citep{hudson2019gqa} & Answer the question using a single word or phrase. \\
    ChartQA \citep{masry2022chartqa} & Answer the question using a single word or phrase. \\
    OCRBench \citep{liu2023ocrbench} & Answer the question using a single word or phrase. \\
    DocVQA \citep{mathew2021docvqa} & Answer the question using a single word or phrase. \\
    InfoVQA \citep{biten2022infovqa} & Answer the question using a single word or phrase. \\
    TextVQA \citep{singh2019towards} & Answer the question using a single word or phrase. \\
    \bottomrule
    \end{tabular}
\end{table}

\begin{table}[t]
    \centering\small
    \caption{
    \textbf{Comparisons on computational costs during the instruction tuning stage} with Cambrian-737K~\citep{tong2024cambrian}, where evaluations are conducted using 8 A100 GPUs with a global batch size of 128.
    Due to the limited GPU memory, we accumulate 4 gradient steps and the batch size per GPU is 4.
    The whole stage requires 5757 training steps.
    GPU memories are averaged over 8 GPUs with DeepSpeed Zero 3.
    }
    \label{tab:cost}
    \vspace{-5pt}
    \setlength{\tabcolsep}{4pt}
    \begin{tabular}{ll c lllll}
    \toprule
    Vision & Base LLM & $\mathcal{L}_{\mathrm{LMM}}^{\mathrm{visual}}$ & \begin{tabular}[c]{@{}l@{}}Trainable\\ Parameters\end{tabular} & Speed (s/iter) & Time & \begin{tabular}[c]{@{}l@{}}GPU\\ Memory\end{tabular} \\
    \midrule
    CLIP-L/336 & Qwen2-7B-Instruct & -- & 7.63 B & 8.31 & 13h 17min & 45.34 G \\
    CLIP-L/336 & Qwen2-7B-Instruct & \checkmark & 7.68 B & 9.02 (1.09$\times$) & 14h 25min & 46.62 G (1.03$\times$) \\

    CLIP-L/336 & Vicuna-13B-v1.5 & -- & 13.05 B & 13.33 & 21h 19min & 48.62 G \\
    CLIP-L/336 & Vicuna-13B-v1.5 & \checkmark & 13.11 B & 14.69 (1.10$\times$) & 23h 30min & 49.07 G (1.01$\times$) \\

    \midrule

    SigLIP-L/384 & Qwen2-7B-Instruct & -- & 7.63 B & 8.77 & 14h 1min & 47.08 G \\
    SigLIP-L/384 & Qwen2-7B-Instruct & \checkmark & 7.68 B & 9.48 (1.08$\times$) & 15h 9min & 52.07 G (1.11$\times$) \\

    SigLIP-L/384 & Vicuna-13B-v1.5 & -- & 13.05 B & 14.22 & 22h 44min & 48.80 G \\
    SigLIP-L/384 & Vicuna-13B-v1.5 & \checkmark & 13.11 B & 15.32 (1.08$\times$) & 24h 30min & 52.68 G (1.08$\times$) \\
    \bottomrule
    \end{tabular}
\end{table}

\begin{figure}[t]
    \centering
    \includegraphics[width=1\linewidth]{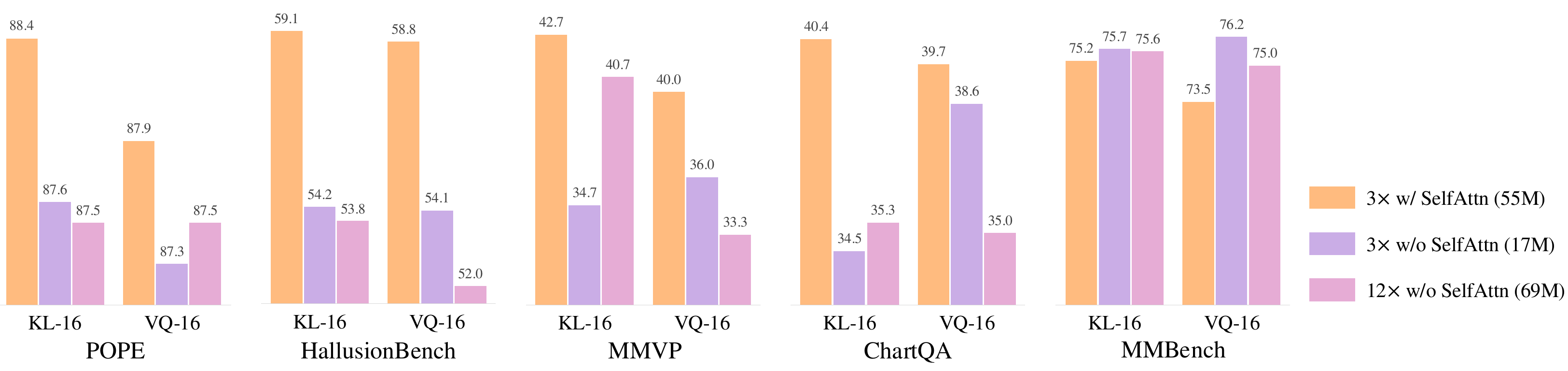}
    \vspace{-18pt}
    \caption{
    \textbf{The architecture of the denoiser and the choice of fine-grained tokenizer.}
    The self-attention module illustrated in \Cref{fig:method}\textcolor{red}{b} is crucial since orange bars consistently outperform others on hallucination and fine-grained comprehension benchmarks, while maintaining similar performances on the general understanding benchmark.
    KL-16 provided by~\cite{rombach2022high} is better than VQ-16 provided by~\cite{sun2024autoregressive}, as quantization may lead to information loss.
    }
    \label{fig:denoising_arch}
\end{figure}

\myparagraph{Hyperparameters.}
The hyperparameters of \method are provided in \Cref{tab:impl}.
We simply borrow most configurations from LLaVA-v1.5~\citep{liu2024improved} without further tuning, as we find it works well with our \method, even if we adopt SigLIP~\citep{zhai2023sigmoid} and Qwen2~\citep{yang2024qwen2} while the original LLaVA-v1.5~\citep{liu2024improved} utilized CLIP~\citep{radford2021learning} and Vicuna-v1.5~\citep{vicuna2023}.
As SigLIP represents a single $384\times384$ image with $729$ tokens and the downsampling ratio of the teacher tokenizer KL-16~\citep{rombach2022high} is $16$, we set the input resolution of the teacher tokenizer as $432 = \sqrt{729}\times16$ to produce $729$ fine-grained tokens as denoising targets.

\myparagraph{Instruction Tuning Data.}
When comparing with state-of-art LMMs in \Cref{tab:main}, our \method is trained on approximately 1.2M instruction tuning data, which is a mixture of Cambrian-737K~\citep{tong2024cambrian} and SMR-473K~\citep{zhang2024beyond}.
Details of these two instruction tuning datasets are listed in \Cref{tab:cambrian} and \Cref{tab:smr}, respectively.
There might be some overlap but we simply concat these two datasets as it is already empirically effective.

\myparagraph{Evaluation Prompts.}
We provide a thorough examination of all evaluation benchmarks utilized in this paper in \Cref{tab:prompt}.
Notably, for MMVP~\citep{tong2024eyes}, which is not officially supported by VLMEvalKit~\citep{duan2024vlmevalkit}, we follow Cambrian-1~\citep{tong2024cambrian} to reformat the original question into a multiple-choice format and compute the accuracy using exact matching.

{\color{textpurple}
\myparagraph{Computational Costs.}
As demonstrated in \Cref{tab:cost}, the denoising process introduces a negligible increase in training time ($\approx$10\% compared to the baseline), while the benefits outweigh the minor additional costs.}

\section{More Experiments}
\label{sec:more_exp}
\vspace{-5pt}
\subsection{More Ablations}\label{sec:more_abl}
\vspace{-5pt}
\myparagraph{KL-16 \textit{v.s.} VQ-16.}
Our default tokenizer is a continuous VAE~\citep{kingma2013auto} with Kullback-Leibler (KL) divergence trained by~\cite{rombach2022high}.
We further conduct experiments with a \textit{discrete} tokenizer provided by~\cite{sun2024autoregressive}, which is a VQGAN~\citep{esser2021taming}, \textit{i.e.}, VQVAE~\citep{oord2017neural} with additional perceptual loss~\citep{zhang2018unreasonable} and adversarial loss~\citep{goodfellow2014generative}.
\textit{KL-16 outperforms VQ-16.}
One intuitive explanation is that KL-16 preserves more low-level details than VQ-16 since quantization may lead to information loss.
Moreover, quantitatively, on ImageNet~\citep{deng2009imagenet} 256$\times$256 validation set, KL-16 achieves 0.87 rFID~\citep{heusel2017gans} while the rFID~\citep{heusel2017gans} of VQ-16 is 2.19.

\myparagraph{Architecture of the Denoiser.}
Illustrated in \Cref{fig:denoising_arch}, \textit{the self-attention module is crucial}, as original visual outputs $\bm{x}_{i \leq N}$ are actually \textit{causal} and we need to model inter-token discrepancy via self-attention.
The number of trainable parameters is \textit{not} the crux.

\begin{table}[t]
    \centering\small
    \caption{
    \textbf{Ablations on different schedules of $\beta$.}
    \method \textit{consistently} improves the baseline, demonstrating its robustness to the denoising schedule.
    }
    \label{tab:schedule}
    \vspace{-5pt}
    \begin{tabular}{l llllll}
    \toprule
    Schedule of $\beta$ & POPE & Hallu. & MMVP & ChartQA & MMB$^{\text{EN}}$ \\
    \midrule
    -- & 87.9 & 55.0 & 29.3 & 34.0 & 73.8 \\
    
    Linear~\citep{ho2020denoising} & 88.1 \up{0.2} & 57.3 \up{2.3} & 42.0 \up{12.4} & 39.2 \up{5.2} & 75.1 \up{1.3} \\
    
    Scaled Linear~\citep{rombach2022high} & \textbf{88.4 \up{0.5}} & 58.3 \up{3.3} & 40.0 \up{10.4} & \textbf{40.7 \up{6.7}} & 75.3 \up{1.5} \\
    
    GLIDE Softmax~\citep{nichol2022glide} & \textbf{88.4 \up{0.5}} & \textbf{59.1 \up{4.1}} & \textbf{42.7 \up{13.4}} & 40.4 \up{6.4} & 75.2 \up{1.4} \\
    
    GeoDiff Sigmoid~\citep{xu2022geodiff} & 88.2 \up{0.3} & 57.7 \up{2.7} & 41.3 \up{11.7} & 38.9 \up{4.9} & \textbf{75.5 \up{1.7}} \\
    \bottomrule
    \end{tabular}
\end{table}

{\color{textpurple}
\myparagraph{Schedule of $\beta$.}
We study the effectiveness of different schedules of $\beta$ in \Cref{tab:schedule}.
From the table, we can tell that even with different schedules of $\beta$, \method \textit{consistently} improves the baseline, demonstrating its robustness to the denoising schedule.
}

\begin{figure}[t]
    \centering
    \includegraphics[width=1\linewidth]{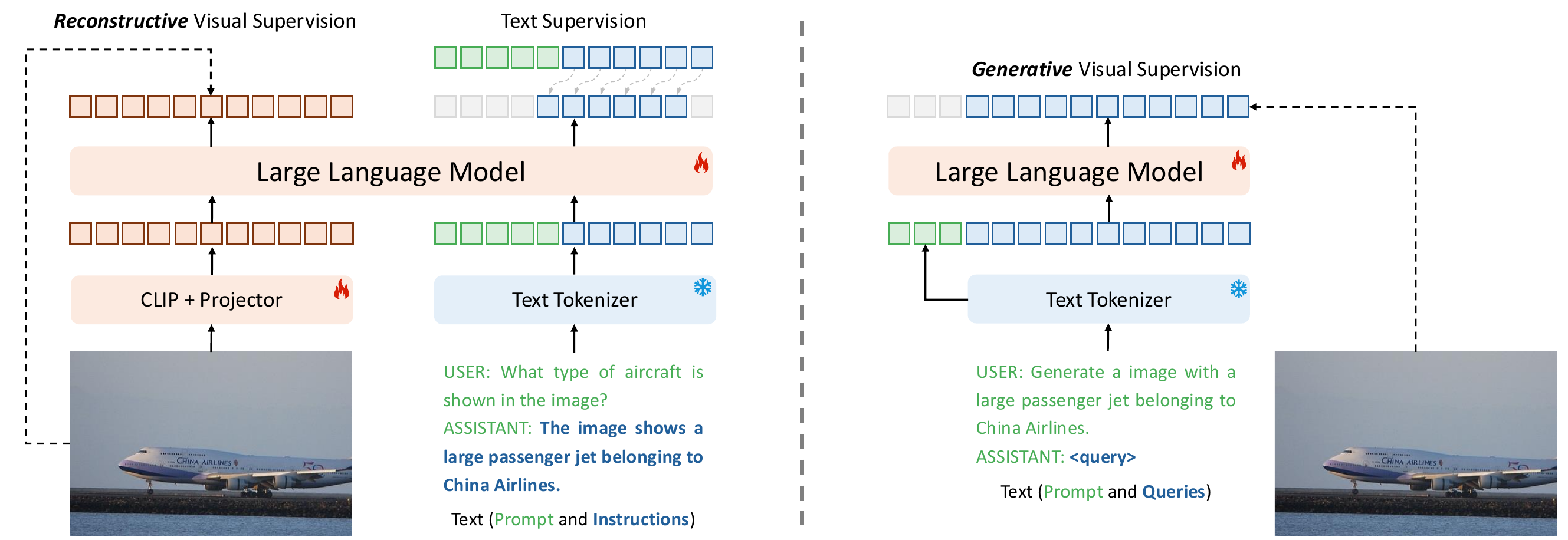}
    \vspace{-15pt}
    \caption{
    \textbf{Pipeline comparison between reconstructive and generative.}
    The reconstructive objective (left) does \textit{not} require specific data formulations and can be easily combined with current visual instruction tuning data.
    However, the generative objective (right) needs specific \textit{text-to-image} creation data, which could be converted by image-to-text caption data.
    }
    \label{fig:gen_rec}
\end{figure}

\myparagraph{Generative \textit{v.s.} Reconstructive.}
We offer a detailed pipeline comparison in \Cref{fig:gen_rec}.
Experimental results have already been provided in \Cref{tab:recon_gen}.
The implementation of generative methods is similar to~\cite{sun2024generative} and~\cite{dong2024dreamllm}, where we adopt 576 learnable queries as inputs and take the corresponding outputs as conditions for the denoiser.

\begin{table}[t]
    \centering\footnotesize
    \caption{
    \textbf{Extended ablations} on The effectiveness of the vision-centric supervision $\mathcal{L}_{\mathrm{LMM}}^{\mathrm{visual}}$ among various LLMs and visual encoders.
    Pre-training data is LLaVA-558K~\citep{liu2023visual} and instruction tuning data is Cambrian-737K~\citep{tong2024cambrian}.
    Evaluations of POPE~\citep{li2023evaluating}, HallusionBench~\citep{guan2024hallusionbench}, MMBench~\citep{liu2023mmbench}, SEED-Bench-1~\citep{li2023seed}, MMMU~\citep{yue2024mmmu}, MMVP \citep{tong2024eyes}, AI2D~\citep{hiippala2021ai2d}, OCRBench~\citep{liu2023ocrbench}, and RealWorldQA~\citep{xai2024grok} are conducted with VLMEvalKit~\citep{duan2024vlmevalkit}, while evaluations of ChartQA~\citep{masry2022chartqa}, DocVQA~\citep{mathew2021docvqa}, InfoVQA~\citep{biten2022infovqa}, and TextVQA~\citep{singh2019towards} are conducted with lmms-eval~\citep{lmmseval2024}.
    }
    \label{tab:extend_abl}
    \vspace{-5pt}
    \setlength{\tabcolsep}{3pt}
    \resizebox{\textwidth}{!}{
    \begin{tabular}{l rr rr rr rr}
    \toprule
    & \multicolumn{4}{c}{CLIP-ViT-L/14@336} & \multicolumn{4}{c}{SigLIP-ViT-SO400M/14@384} \\
    \cmidrule(lr){2-5} \cmidrule(lr){6-9}
    & \multicolumn{2}{c}{Vicuna-7B-v1.5} & \multicolumn{2}{c}{Qwen2-7B-Instruct} & \multicolumn{2}{c}{Vicuna-7B-v1.5} & \multicolumn{2}{c}{Qwen2-7B-Instruct} \\
    \cmidrule(lr){2-3} \cmidrule(lr){4-5} \cmidrule(lr){6-7} \cmidrule(lr){8-9}
    Benchmark & LLaVA & \multicolumn{1}{c}{\method} & LLaVA & \multicolumn{1}{c}{\method} & LLaVA & \multicolumn{1}{c}{\method} & LLaVA & \multicolumn{1}{c}{\method} \\
    \midrule
    
    POPE$_{\text{acc}}$ & 86.3 & \textbf{87.2 \up{0.9}} & 87.9 & \textbf{88.4 \up{0.5}} & 86.0 & \textbf{87.7 \up{1.7}} & 88.5 & \textbf{88.7 \up{0.2}} \\
    
    HallusionBench$_{\text{aAcc}}$ & 52.5 & \textbf{55.8 \up{3.3}} & 55.0 & \textbf{59.1 \up{4.1}} & 50.4 & \textbf{53.8 \up{3.4}} & 57.3 & \textbf{58.2 \up{0.9}} \\
    
    MMBench-EN$_{\text{dev}}$ & 67.0 & \textbf{67.6 \up{0.6}} & 73.8 & \textbf{75.2 \up{1.4}} & 64.5 & \textbf{69.2 \up{4.7}} & 76.3 & \textbf{76.9 \up{0.6}} \\
    
    MMBench-CN$_{\text{dev}}$ & \textbf{60.0} & 59.8 \down{0.2} & 72.9 & \textbf{73.7 \up{0.8}} & 63.1 & \textbf{63.4 \up{0.3}} & 75.7 & \textbf{76.3 \up{0.7}} \\
    
    SEED$_{\text{img}}$ & \textbf{66.7} & 66.4 \down{0.3} & 70.3 & \textbf{70.7 \up{0.4}} & 68.2 & \textbf{69.0 \up{0.8}} & \textbf{72.3} & 72.1 \down{0.2} \\
    
    MMMU$_{\text{dev}}$ & 30.0 & \textbf{34.0 \up{4.0}} & 44.0 & \textbf{45.3 \up{1.3}} & 33.3 & \textbf{38.0 \up{4.7}} & 38.7 & \textbf{41.3 \up{2.6}} \\
    
    MMMU$_{\text{val}}$ & 35.3 & \textbf{36.0 \up{0.7}} & 41.9 & \textbf{42.6 \up{0.7}} & 34.2 & \textbf{35.4 \up{1.2}} & 41.8 & \textbf{43.8 \up{2.0}} \\
    
    MMVP & 28.0 & \textbf{36.0 \up{8.0}} & 29.3 & \textbf{42.7 \up{13.4}} & 27.3 & \textbf{38.0 \up{10.7}} & 40.7 & \textbf{49.3 \up{8.6}} \\
    
    AI2D$_{\text{test}}$ & 61.2 & \textbf{61.4 \up{0.2}} & 71.9 & \textbf{73.3 \up{1.4}} & \textbf{62.6} & 62.4 \down{0.2} & 74.0 & \textbf{74.5 \up{0.5}} \\
    
    ChartQA$_{\text{test}}$ & 32.9 & \textbf{39.8 \up{6.9}} & 36.2 & \textbf{41.6 \up{5.4}} & 34.0 & \textbf{48.2 \up{14.2}} & 44.4 & \textbf{46.9 \up{2.5}} \\
    
    DocVQA$_{\text{val}}$ & 33.4 & \textbf{41.6 \up{8.2}} & 31.1 & \textbf{44.7 \up{13.6}} & 40.4 & \textbf{40.7 \up{0.3}} & 39.2 & \textbf{39.3 \up{0.1}} \\
    
    InfoVQA$_{\text{val}}$ & 21.2 & \textbf{26.4 \up{5.2}} & 22.1 & \textbf{39.3 \up{16.2}} & 22.8 & \textbf{23.3 \up{0.5}} & 24.0 & \textbf{25.1 \up{1.1}} \\
    
    TextVQA$_{\text{val}}$ & 55.7 & \textbf{58.7 \up{3.0}} & 52.0 & \textbf{54.1 \up{2.1}} & 60.5 & \textbf{62.6 \up{2.1}} & 56.3 & \textbf{57.5 \up{1.2}} \\
    
    OCRBench & 339 & \textbf{350 \up{11}} & 363 & \textbf{381 \up{18}} & 354 & \textbf{365 \up{11}} & 432 & \textbf{448 \up{16}} \\
    
    RealWorldQA & 52.7 & \textbf{53.2 \up{0.5}} & 56.7 & \textbf{57.4 \up{0.7}} & 55.0 & \textbf{57.1 \up{2.1}} & 57.9 & \textbf{59.1 \up{1.2}} \\
    
    \cmidrule{1-9}

    Average & 47.8 & \textbf{50.6 \up{2.8}} & 52.1 & \textbf{56.4 \up{4.3}} & 49.2 & \textbf{52.4 \up{3.2}} & 55.4 & \textbf{56.9 \up{1.5}} \\
    
    \bottomrule
    \end{tabular}}
    \vspace{-10pt}
\end{table}

{
\color{textpurple}

We hypothesize that the underlying reason for the lower performance of generative methods in comprehension tasks is \textit{the weak correspondence between inputs and supervision} under generative settings, which typically arises from both the (1) data and the (2) design of these methods.

(1) Typical generative methods that explore the synergy of comprehension and generation, usually leverage image generation conditioned on text instructions on \textit{(i) text-to-image datasets} or \textit{(ii) interleaved datasets} as extra supervision.
However, (i) text-to-image datasets are typically designed to generate \textit{high-aesthetic} samples rather than text-aligned ones, and (ii) interleaved datasets aim to enable few-shot learning via interleaving independent supervised examples, where reasoning becomes more important than alignment.
Therefore, there exists a clear disconnect where the supervision (image) has little to do with the input (text instruction).
For example, the CLIP-Score~\citep{hessel2021clipscore}, which measures the similarity between text and images, is only 0.3043 for the LAION-Art dataset~\citep{schuhmann2022laion} and 0.2842 for the MMC4 dataset~\citep{zhu2023multimodal}, indicating that the supervision signals in these datasets are \textit{not} well-suited for tasks requiring strong text-image alignment.

(2) Even when we attempt to ensure image-text alignment by converting aligned caption data into creation data for supervision, the results demonstrated in \Cref{tab:recon_gen} remain unsatisfactory.
This suggests that the \textit{design of generative objectives itself does not inherently require a strong correspondence} between inputs and supervision targets.

In contrast, reconstructive methods like \method leverage the original input images as auxiliary supervision, ensuring a strong and direct correspondence, which is crucial for tasks requiring accurate comprehension and interpretation of multimodal data, leading to significantly improved performance.

\myparagraph{Extended Ablations on Different LLMs and Visual Encoders.}
We extend the ablation in \Cref{tab:llm} by incorporating more benchmarks, providing a more balanced and representative distribution of tasks.
Empirical results in \Cref{tab:extend_abl} demonstrate that our proposed vision-centric supervision utilized by \method leads to significant improvements in most cases.
Moreover, we found \method contributes more significant improvements over fine-grained comprehension datasets, such as HallusionBench~\citep{guan2024hallusionbench}, MMVP~\citep{tong2024eyes}, ChartQA~\citep{masry2022chartqa}, and OCRBench~\citep{liu2023ocrbench}.
}

{
\color{textpurple}

\subsection{Comparison on High-Resolution Benchmarks}\label{sec:ocr}
\vspace{-5pt}
We incorporate the ``anyres'' technique proposed by LLaVA-v1.6~\citep{liu2024llavanext} into our \method.
Specifically, for each image, we employ a grid configuration of 384$\times$\{2$\times$2, 1$\times$\{2,3,4\}, \{2,3,4\}$\times$1\} to identify the input resolution, resulting in a maximum of 5$\times$729 $=$ 3,645 visual tokens.
\textit{Each} 384$\times$384 crop is required to reconstruct the original input via the denoising objective proposed by \method.
In \Cref{tab:ocr}, our \textbf{\method-7B}$_{\text{anyres}}$ surpasses LLaVA-v1.6-7B~\citep{liu2024llavanext} and Cambrian-1-8B~\citep{tong2024cambrian} under most cases.
These results indicate that \method not only performs well at lower resolutions but also maintains its competitive edge at higher resolutions, making it a robust and versatile method.
}

\begin{table}[t]
    \centering\small
    \caption{
    \textbf{Comparison to state-of-the-art LMMs on benchmarks requires high-resolution inputs.}
    We evaluate models on: ChartQA~\citep{masry2022chartqa}, DocVQA~\citep{mathew2021docvqa} val set, InfoVQA~\citep{biten2022infovqa} val set, TextVQA~\citep{singh2019towards} val set, OCRBench~\citep{liu2023ocrbench}, and RealWorldQA~\citep{xai2024grok}.
    %
    %
    $^\ddag$We evaluate the official checkpoint.
    }
    \vspace{-5pt}
    \setlength{\tabcolsep}{3pt}
    \resizebox{\textwidth}{!}{
    \begin{tabular}{r cccccc}
        \toprule
        Model & ChartQA & DocVQA & InfoVQA & TextVQA  & OCRBench & RealWorldQA\\
        \midrule

        GPT-4V-1106 \citep{gpt4v}  & 78.5 & 88.4 & -- & 78.0 & 645 & 61.4 \\
        Gemini-1.5 Pro \citep{team2023gemini} & 81.3 & 86.5 & -- & 78.1 & -- & 67.5 \\
        Grok-1.5 \citep{xai2024grok} & 76.1 & 85.6 & -- & 78.1 & -- & 68.7 \\
            
        \midrule 
        
        LLaVA-v1.5-7B$^\ddag$ \citep{liu2024improved} & 18.2 & 28.1 & 25.7 & 58.2 & 317 & 54.9 \\
        LLaVA-v1.6-7B$^\ddag$ \citep{liu2024llavanext} & 65.5 &  74.4 & 37.1 & 64.8 & 532 & 57.6 \\
        Cambrian-1-8B \citep{tong2024cambrian} & 73.3 & 77.8 & -- & 71.7 & \textbf{624} & 64.2 \\
        
        \textbf{\method-7B}$_{\text{anyres}}$ & \textbf{76.9} & \textbf{81.8} & \textbf{50.5} & \textbf{72.2} & 607 & \textbf{66.2} \\
        \bottomrule
    \end{tabular}
    }
    \label{tab:ocr}
\end{table}

\begin{table}[t]
    \centering\small
    \caption{
    \textbf{Evaluations on language performance.}
    We evaluate multi-modal benchmarks that mainly require general knowledge following~\cite{tong2024cambrian}.
    Furthermore, we incorporate representative language benchmarks, including general understanding on MMLU~\citep{hendrycks2020measuring} and HellaSwag~\citep{zellers2019hellaswag}, and instruction-following on IFEval~\citep{zhou2023instruction}.
    \method does \textit{not} harm language capabilities as it brings improvements in most cases.
    }
    \label{tab:language}
    \vspace{-5pt}
    \setlength{\tabcolsep}{4pt}
    \begin{tabular}{l rr rr rr rr}
    \toprule
    & \multicolumn{4}{c}{CLIP-ViT-L/14@336} & \multicolumn{4}{c}{SigLIP-ViT-SO400M/14@384} \\
    \cmidrule(lr){2-5} \cmidrule(lr){6-9}
    
    & \multicolumn{2}{c}{Vicuna-7B-v1.5} & \multicolumn{2}{c}{Qwen2-7B-Instruct} & \multicolumn{2}{c}{Vicuna-7B-v1.5} & \multicolumn{2}{c}{Qwen2-7B-Instruct} \\
    
    \cmidrule(lr){2-3} \cmidrule(lr){4-5} \cmidrule(lr){6-7} \cmidrule(lr){8-9}
    Benchmark & LLaVA & \multicolumn{1}{c}{\method} & LLaVA & \multicolumn{1}{c}{\method} & LLaVA & \multicolumn{1}{c}{\method} & LLaVA & \multicolumn{1}{c}{\method} \\
    
    \hline
    \rowcolor{lightgray}
    \multicolumn{9}{l}{\textit{Vision-Language Benchmarks on Knowledge}} \\

    ScienceQA$_{\text{test}}$ & 68.5 & \textbf{69.0 \up{0.5}} & 76.5 & \textbf{77.4 \up{0.9}} & 69.6 & \textbf{71.3 \up{1.7}} & 78.3 & \textbf{78.5 \up{0.2}} \\
    
    MMMU$_{\text{dev}}$ & 30.0 & \textbf{34.0 \up{4.0}} & 44.0 & \textbf{45.3 \up{1.3}} & 33.3 & \textbf{38.0 \up{4.7}} & 38.7 & \textbf{41.3 \up{2.6}} \\
    
    MMMU$_{\text{val}}$ & 35.3 & \textbf{36.0 \up{0.7}} & 41.9 & \textbf{42.6 \up{0.7}} & 34.2 & \textbf{35.4 \up{1.2}} & 41.8 & \textbf{43.8 \up{2.0}} \\
    
    AI2D$_{\text{test}}$ & 61.2 & \textbf{61.4 \up{0.2}} & 71.9 & \textbf{73.3 \up{1.4}} & \textbf{62.6} & 62.4 \down{0.2} & 74.0 & \textbf{74.5 \up{0.5}} \\
    
    \hline
    \rowcolor{lightgray}
    \multicolumn{9}{l}{\textit{Language Benchmarks}} \\
    
    MMLU & 26.5 & \textbf{27.4 \up{0.9}} & 57.1 & \textbf{60.7 \up{3.6}} & \textbf{26.0} & 25.9 \down{0.1} & 60.9 & \textbf{61.0 \up{0.1}} \\
    
    HellaSwag$_{\text{acc-norm}}$ & \textbf{27.0} & 26.9 \down{0.1} & \textbf{46.4} & 46.2 \down{0.2} & \textbf{27.1} & 27.0 \down{0.1} & 45.5 & \textbf{46.6 \up{1.1}} \\
    
    IFEval$_{\text{strict-inst}}$ & 41.2 & \textbf{44.6 \up{3.4}} & 47.1 & \textbf{49.2 \up{2.1}} & 43.6 & \textbf{43.8 \up{0.2}} & 47.8 & \textbf{48.1 \up{0.3}} \\
    
    IFEval$_{\text{strict-prompt}}$ & 28.7 & \textbf{35.3 \up{6.7}} & 35.1 & \textbf{37.0 \up{1.9}}  & 32.5 & \textbf{33.1 \up{0.6}} & 35.3 & \textbf{36.2 \up{0.9}} \\

    \cmidrule{1-9}
    Average & 39.8 & \textbf{41.8 \up{2.0}} & 52.5 & \textbf{54.0 \up{1.5}} & 41.1 & \textbf{42.1 \up{1.0}} & 52.8 & \textbf{53.8 \up{1.0}} \\
    
    \bottomrule
    \end{tabular}
    \vspace{-10pt}
\end{table}

\begin{table}[t]
    \centering\footnotesize
    \caption{
    \textbf{Model scaling of \method.}
    We take Qwen2.5 series~\citep{qwen2.5} as the base language model and CLIP-ViT-L/14@336~\citep{radford2021learning} as the visual encoder.
    Pre-training data is LLaVA-558K~\citep{liu2023visual} and the instruction tuning data is LLaVA-665K~\citep{liu2024improved}.
    \textit{\method brings improvements over the baseline across different model sizes in most cases}.
    }
    \label{tab:model_scale}
    \vspace{-5pt}
    \setlength{\tabcolsep}{3.5pt}
    \begin{tabular}{l rr rr rr rr}
    \toprule
    \multirow{2}{*}{Benchmark} & \multicolumn{2}{c}{0.5B} & \multicolumn{2}{c}{1.5B} & \multicolumn{2}{c}{3B} & \multicolumn{2}{c}{7B} \\
    \cmidrule(lr){2-3} \cmidrule(lr){4-5} \cmidrule(lr){6-7} \cmidrule(lr){8-9}
    & LLaVA & \multicolumn{1}{c}{\method} & LLaVA & \multicolumn{1}{c}{\method} & LLaVA & \multicolumn{1}{c}{\method} & LLaVA & \multicolumn{1}{c}{\method} \\
    \midrule
    
    POPE$_{\text{acc}}$ & 50.0 & \textbf{60.4 \up{10.4}} & 85.3 & \textbf{87.9 \up{2.4}} & 87.3 & \textbf{88.1 \up{0.8}} & 87.9 & \textbf{88.4 \up{0.5}} \\
    
    HallusionBench$_{\text{aAcc}}$ & 45.8 & \textbf{48.0 \up{2.2}} & 48.7 & \textbf{49.6 \up{0.9}} & 52.2 & 52.2 \textcolor{mygray}{-- 0.0} & 48.7 & \textbf{53.7 \up{5.0}} \\
    
    MMBench-EN$_{\text{dev}}$ & 55.2 & \textbf{60.4 \up{5.2}} & 67.5 & \textbf{68.2 \up{1.7}} & 70.6 & \textbf{71.4 \up{0.8}} & 75.0 & \textbf{75.7 \up{0.7}} \\
    
    MMBench-CN$_{\text{dev}}$ & 45.6 & \textbf{48.9 \up{3.3}} & 62.4 & \textbf{63.9 \up{1.5}} & 68.0 & \textbf{69.1 \up{1.1}} & \textbf{73.6} & 73.5 \down{0.1} \\
    
    SEED$_{\text{img}}$ & \textbf{55.8} & 55.6 \down{0.2} & 66.3 & \textbf{66.8 \up{0.5}} & 68.2 & \textbf{68.4 \up{0.2}} & 70.6 & \textbf{71.0 \up{0.4}} \\
    
    OCRBench & 229 & \textbf{248 \up{19}} & 291 & \textbf{298 \up{7}} & \textbf{313} & 308 \down{5} & 334 & \textbf{358 \up{24}} \\
    
    MMMU$_{\text{dev}}$ & 35.2 & \textbf{36.0 \up{0.8}} & 44.7 & \textbf{45.0 \up{0.3}} & 48.7 & \textbf{49.0 \up{0.3}} & 48.0 & 48.0 \textcolor{mygray}{-- 0.0} \\
    
    MMMU$_{\text{val}}$ & 38.0 & \textbf{40.3 \up{1.7}} & 41.8 & \textbf{43.6 \up{1.8}} & 41.6 & \textbf{42.7 \up{1.1}} & 47.3 & \textbf{48.0 \up{0.7}} \\
    
    AI2D$_{\text{test}}$ & 45.3 & \textbf{46.0 \up{0.7}} & 59.0 & \textbf{59.5 \up{0.5}} & 62.9 & \textbf{63.2 \up{0.3}} & 68.3 & \textbf{68.5 \up{0.2}} \\
    
    RealWorldQA & 45.1 & \textbf{46.4 \up{1.3}} & 50.5 & \textbf{53.5 \up{3.0}} & 55.7 & \textbf{57.9 \up{2.2}} & 59.5 & \textbf{59.9 \up{0.4}} \\

    \cmidrule{1-9}
    Average & 43.9 & \textbf{46.7 \up{2.8}} & 55.3 & \textbf{56.8 \up{1.5}} & 58.9 & \textbf{59.3 \up{0.4}} & 61.2 & \textbf{62.3 \up{1.1}} \\
    
    \bottomrule
    \end{tabular}

    \vspace{10pt}

    \centering\small
    \caption{
    \textbf{Data scaling of \method.}
    We take Qwen2-7B-Instruct~\citep{yang2024qwen2} as the base language model and CLIP-ViT-L/14@336~\citep{radford2021learning} as the visual encoder.
    \textit{\method consistently brings significant improvements as the training data scale increases.}
    }
    \label{tab:data_scale}
    \vspace{-5pt}
    \begin{tabular}{llc llllll}
    \toprule
    PT & SFT & $\mathcal{L}_{\mathrm{LMM}}^{\mathrm{visual}}$ & POPE & Hallu. & ChartQA & OCRBench & MMB$^{\text{EN}}$ & AI2D \\
    
    \midrule
    \multirow{2}{*}{558K} & \multirow{2}{*}{737K} & -- & 87.9 & 55.0 & 34.0 & 363 & 73.8 & 72.4 \\
    & & \checkmark & \textbf{88.4 \up{0.5}} & \textbf{59.1 \up{4.1}} & \textbf{40.4 \up{6.4}} & \textbf{380 \up{17}} & \textbf{75.2 \up{1.4}} & \textbf{73.3 \up{0.9}} \\

    \midrule
    \multirow{2}{*}{558K} & \multirow{2}{*}{1.2M} & -- & 88.5 & 57.3 & 37.0 & 389 & 75.7 & 74.5 \\
    & & \checkmark & \textbf{88.8 \up{0.3}} & \textbf{57.8 \up{0.5}} & \textbf{42.0 \up{5.0}} & \textbf{392 \up{3}} & \textbf{76.8 \up{1.1}} & \textbf{74.7 \up{0.2}} \\

    \midrule
    \multirow{2}{*}{2M} & \multirow{2}{*}{737K} & -- & 88.1 & 55.6 & 37.3 & 384 & 76.2 & 72.3 \\
    & & \checkmark & \textbf{88.3 \up{0.2}} & \textbf{56.2 \up{0.6}} & \textbf{41.9 \up{4.5}} & \textbf{398 \up{14}} & \textbf{77.0 \up{0.8}} & \textbf{73.4 \up{1.1}} \\

    \midrule
    \multirow{2}{*}{2M} & \multirow{2}{*}{1.2M} & -- & 88.5 & 53.8 & 41.2 & 388 & 76.5 & 73.9 \\
    & & \checkmark & \textbf{88.9 \up{0.4}} & \textbf{57.3 \up{2.5}} & \textbf{43.2 \up{2.0}} & \textbf{405 \up{17}} & \textbf{78.0 \up{1.5}} & \textbf{74.1 \up{0.2}} \\
    
    \bottomrule
    \end{tabular}
\end{table}

\begin{figure}[t]
    \centering
    \includegraphics[width=0.85\linewidth]{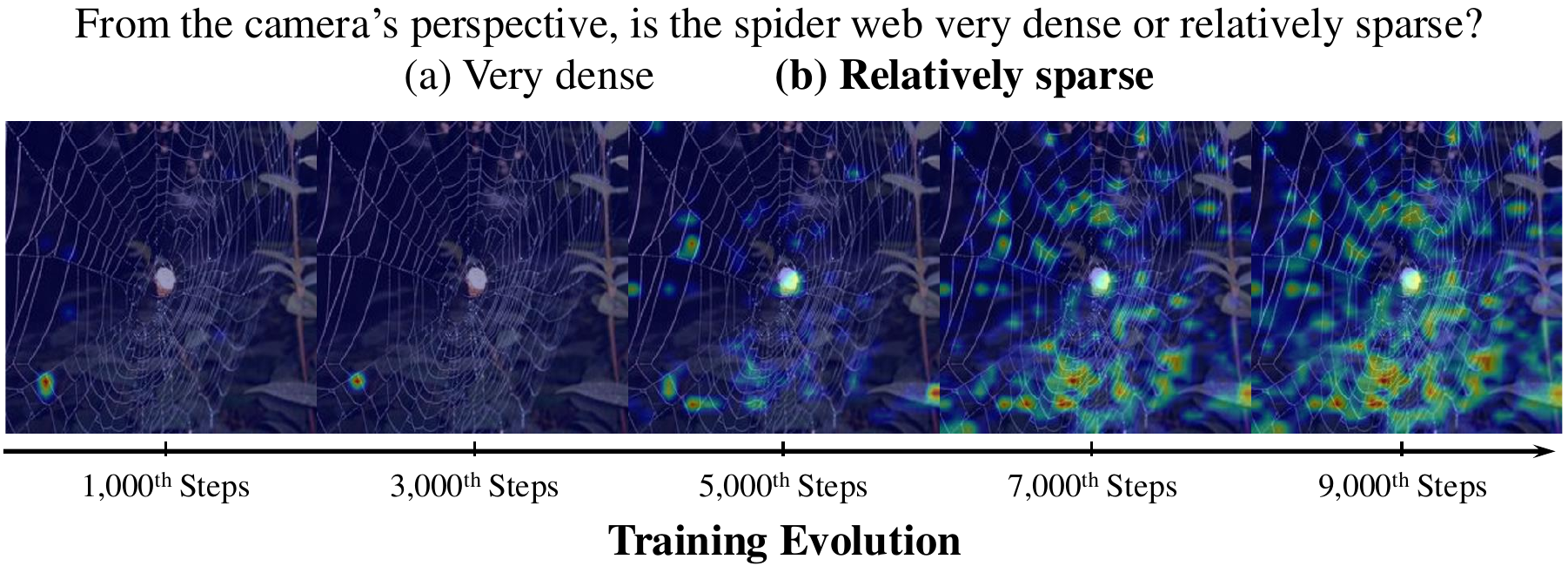}
    \vspace{-5pt}
    \caption{
    \textbf{Qualitative comparison using GradCAM~\citep{selvaraju2020grad}} on MMVP~\citep{tong2024eyes}.
    We visualize the gradient of the second-to-last block of the LMM using the option of the ground-truth answer as the target class.
    Equipped with our proposed vision-centric supervision signals, \method provides more reasonable gradients and urges LMMs to focus on relevant regions (the spider web) as the training evolves.
    }
    \label{fig:gradcam}
    \vspace{-10pt}
\end{figure}

{
\color{textpurple}

\subsection{More Analysis}\label{sec:more_ana}
\vspace{-5pt}
\myparagraph{Language Capabilities.}
One possible concern of \method is that this type of low-level reconstruction may harm the high-level language capabilities.
To investigate this issue, we evaluate multi-modal benchmarks that mainly require general knowledge following~\citep{tong2024cambrian}, including ScienceQA~\citep{saikh2022scienceqa}, MMMU~\citep{yue2024mmmu}, and AI2D~\citep{hiippala2021ai2d}.
Furthermore, we incorporate representative language benchmarks, including general understanding on MMLU~\citep{hendrycks2020measuring} and HellaSwag~\citep{zellers2019hellaswag}, and instruction-following on IFEval~\citep{zhou2023instruction}.
Empirical results in \Cref{tab:language} demonstrate that \method does \textit{not} harm language capabilities as it brings improvements in most cases.

\begin{figure}[t]
    \centering
    \includegraphics[width=1\linewidth]{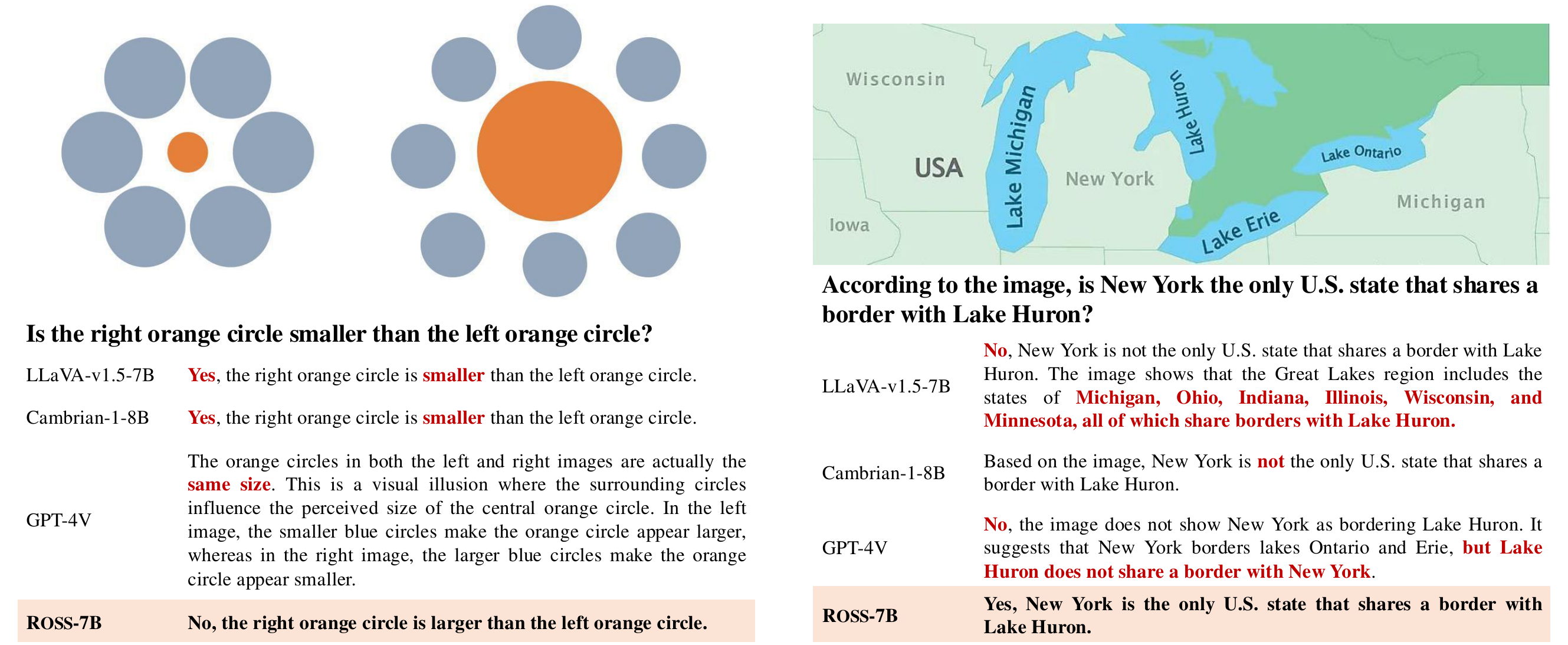}
    \vspace{-15pt}
    \caption{
    \textbf{Qualitative comparisons on HallusionBench~\citep{guan2024hallusionbench}.}
    }
    \label{fig:hallusion}
\end{figure}

\begin{figure}[t]
    \includegraphics[width=1\linewidth]{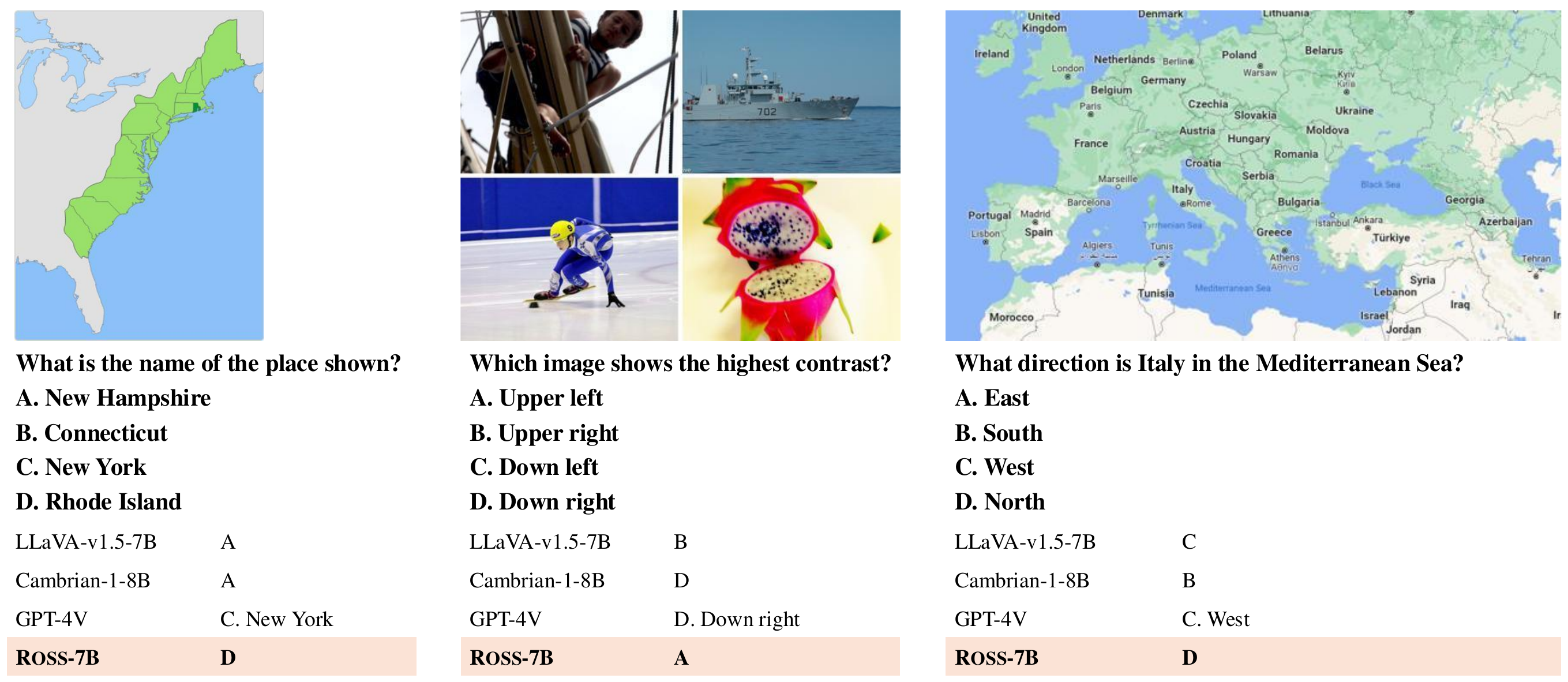}
    \vspace{-15pt}
    \caption{
    \textbf{Qualitative comparisons on MMbench~\citep{guan2024hallusionbench} English dev split.}
    }
    \label{fig:mmbench}
    \vspace{-10pt}
\end{figure}

\myparagraph{Model Scaling Properties.}
To study the stability and scalability of \method across different model sizes, we use the Qwen2.5 series~\citep{qwen2.5} with varying sizes as the base language model while keeping the CLIP-ViT-L/14@336~\citep{radford2021learning} as the visual encoder.
The pre-training data is LLaVA-558K~\citep{liu2023visual}, and the instruction tuning data is LLaVA-665K~\citep{liu2024improved}.
The results, shown in \Cref{tab:model_scale}, demonstrate that \textit{\method brings improvements over the baseline (LLaVA) across different model sizes in most cases}.

\myparagraph{Data Scaling Properties.}
To study the impact of the training data scale, we used Qwen2-7B-Instruct~\citep{yang2024qwen2} as the base language model and CLIP-ViT-L/14@336~\citep{radford2021learning} as the visual encoder.
We compared the performance of \method and the baseline under different scales of training data.
\Cref{tab:data_scale} demonstrates that \textit{\method consistently brings significant improvements as the training data scale increases.}

\myparagraph{Gradient Analysis.}
To better explain the reasoning behind how the vison-centric supervision enables the model to focus on relevant areas of the image during VQA tasks, we provide qualitative comparison using GradCAM~\citep{selvaraju2020grad} on MMVP~\citep{tong2024eyes} in \Cref{fig:gradcam}, since GradCAM helps in understanding which parts of the image the model is focusing on, making the model's decision-making process more transparent.
In our analysis, we visualize the gradient of the second-to-last block of the LMM, regarding the option of the ground-truth answer as the target class.
Specifically in this case, where the providing question is about the spider web, our proposed vision-centric supervision signals provide more reasonable gradients and urge LMMs to focus on relevant regions, \textit{i.e.}, the spider web, as the training evolves.

}

\begin{figure}[t]
    \includegraphics[width=1\linewidth]{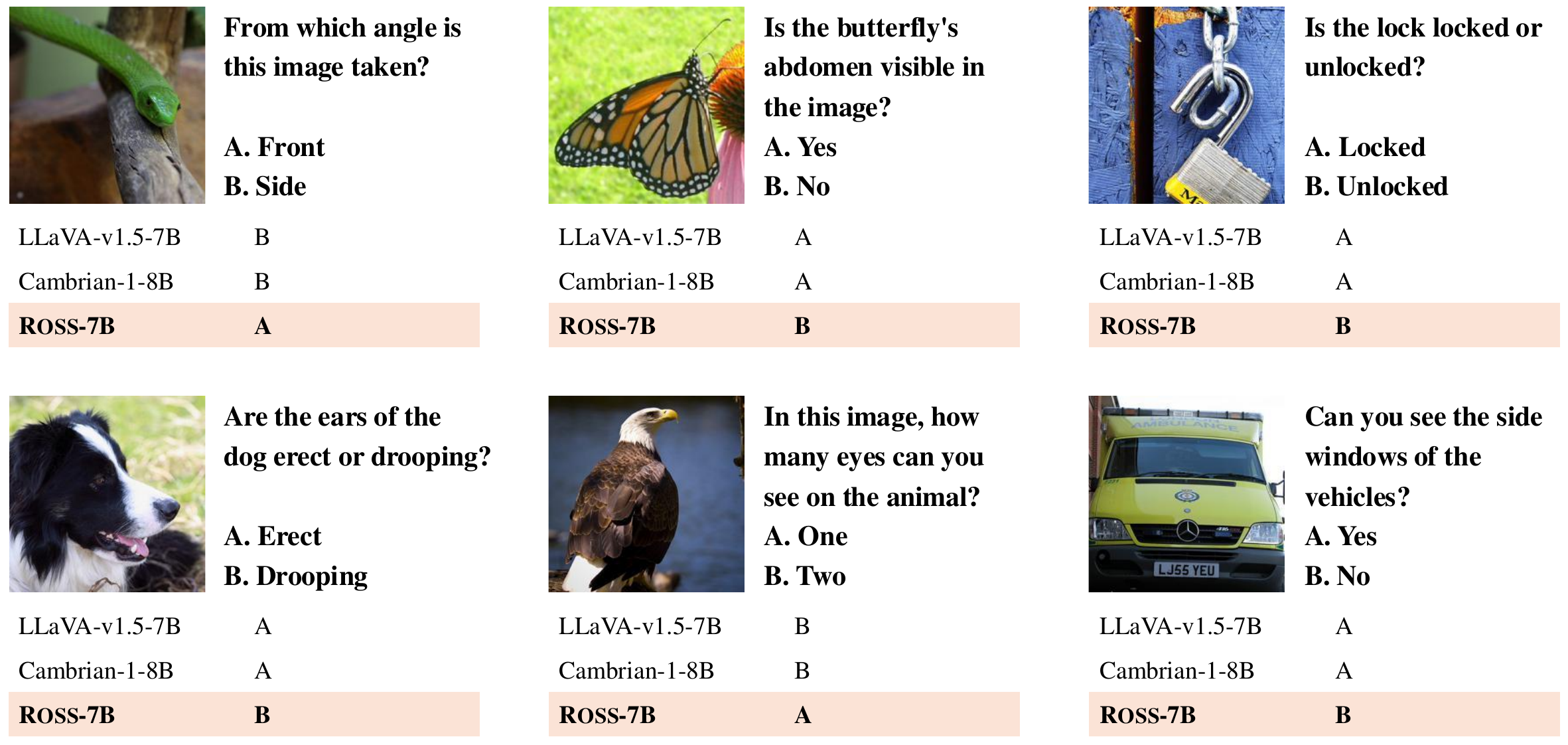}
    \vspace{-15pt}
    \caption{
    \textbf{Qualitative comparisons on MMVP~\citep{tong2024eyes}.}
    }
    \label{fig:mmvp}
\end{figure}

\begin{figure}[t]
    \centering
    \includegraphics[width=1\linewidth]{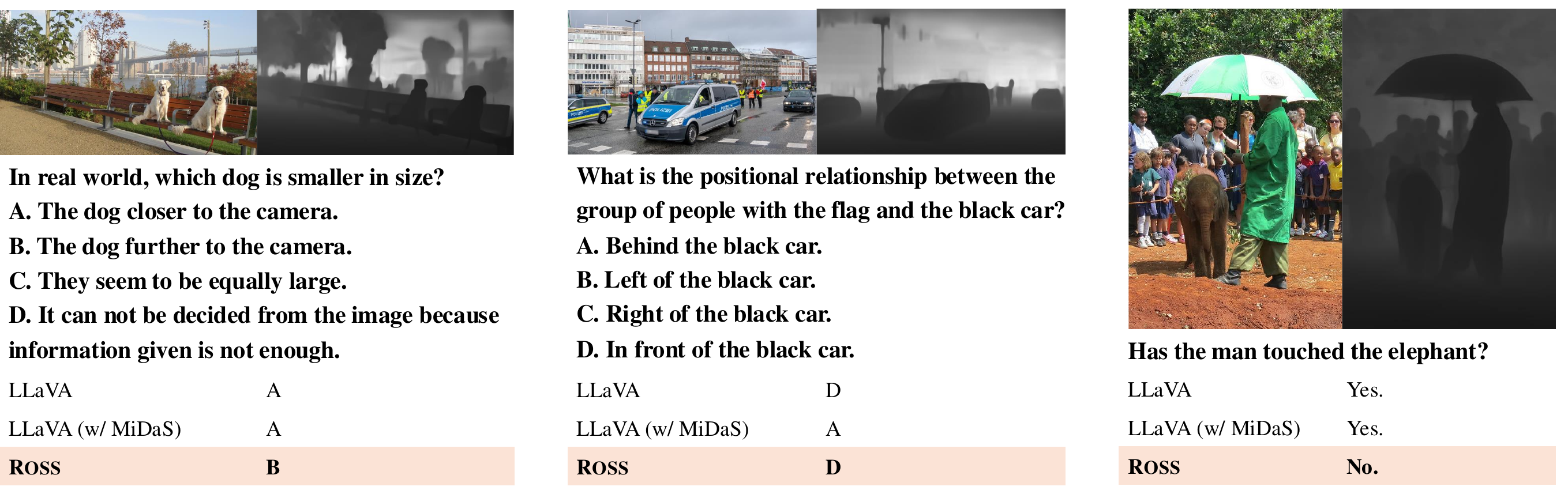}
    \vspace{-15pt}
    \caption{
    \textbf{Qualitative comparisons} on SpatialBench~\citep{cai2024spatialbot}.
    We take RGB + D inputs when testing.
    %
    %
    Notably, the extra depth expert MiDaS-3.0~\citep{birkl2023midas} sometimes \textit{harms} comprehension (see the second example).
    }
    \label{fig:spatialbench}
    \vspace{-10pt}
\end{figure}

\subsection{Qualitative Comparisons}\label{sec:visual}
\vspace{-5pt}
We provide sufficient qualitative comparisons in \Cref{fig:hallusion}, \Cref{fig:mmbench}, \Cref{fig:mmvp}, and \Cref{fig:spatialbench} on HallusionBench~\citep{guan2024hallusionbench}, MMBench~\citep{liu2023mmbench} English dev split, MMVP~\citep{tong2024eyes}, and SpatialBench~\citep{cai2024spatialbot}, respectively.
In \Cref{fig:hallusion}, \Cref{fig:mmbench}, and \Cref{fig:mmvp}, we compare our \textbf{\method-7B} with the instruction tuning baseline LLaVA-v1.5-7B~\citep{liu2024improved}, the state-of-the-art open-source method using extrinsic assistance Cambrian-1-8B~\citep{tong2024cambrian}, and GPT-4V~\citep{gpt4v}.

As demonstrated in \Cref{fig:hallusion}, where we highlight the \textit{wrong} parts of each prediction in red, our \method manages to correctly answer the question with reduced hallucinations even when GPT-4V fails.
Cambrian-1~\citep{tong2024cambrian} even fails to follow the instructions in the second example.
This could be because a super huge SFT data (7M) may harm the instruction-following abilities of LMMs.
Qualitative results shown in \Cref{fig:mmbench} demonstrate both enhanced reasoning abilities (the first example), low-level comprehension capabilities (the second example), and spatial understanding skills (the third example).
\Cref{fig:mmvp} illustrates that our \method is good at recognizing various visual patterns, implying that the introduced reconstructive vision-centric objective indeed makes up the visual shortcomings of the original visual encoder.

\Cref{fig:spatialbench} provides qualitative results on SpatialBench~\citep{cai2024spatialbot}.
The extra depth understanding visual expert, \textit{i.e.}, MiDaS~\citep{birkl2023midas}, fails to help LMMs understand depth maps both quantitatively in \Cref{tab:depth} and qualitatively in \Cref{fig:spatialbench}.

\section{Discussion}
\vspace{-5pt}
One limitation is that \method does not have generation capabilities, since \method is designed for enhanced multimodal comprehension, without the need to generate photorealistic aesthetic images.
Furthermore, the gap in \textit{training data} between comprehension and generation methods also matters.
For instance, PixArt-$\alpha$~\citep{chen2023pixart}, which is one of the most \textit{efficient} text-to-image models, was trained on nearly 400M images to model the pixel discrepancy just in the \textit{first} training stage.
By contrast, our \method is only trained on nearly 3M images for one epoch.
Future topics include achieving photorealistic text-to-image generation via incorporating more training samples.

\end{document}